\documentclass[sigconf]{acmart}
\usepackage{amsmath}
\usepackage[ruled,linesnumbered]{algorithm2e}
\usepackage{bbm}
\usepackage{bm}
\usepackage{multirow}
\usepackage{makecell} 
\usepackage{tabto}
\usepackage{graphicx}
\usepackage{float} 
\usepackage{subfigure}
\usepackage{booktabs}
\usepackage{enumitem}
\newtheorem{myDef}{\textbf{Definition}}

\newtheorem{Proposition}{\textbf{Proposition}}
\usepackage{hyperref}
\usepackage[hyphenbreaks]{breakurl}

\allowdisplaybreaks[4]

\AtBeginDocument{%
  \providecommand\BibTeX{{%
    \normalfont B\kern-0.5em{\scshape i\kern-0.25em b}\kern-0.8em\TeX}}}

\setcopyright{acmcopyright}
\copyrightyear{2018}
\acmYear{2018}
\acmDOI{10.1145/1122445.1122456}

\acmConference[Woodstock '18]{Woodstock '18: ACM Symposium on Neural
  Gaze Detection}{June 03--05, 2018}{Woodstock, NY}
\acmBooktitle{Woodstock '18: ACM Symposium on Neural Gaze Detection,
  June 03--05, 2018, Woodstock, NY}
\acmPrice{15.00}
\acmISBN{978-1-4503-XXXX-X/18/06}

\begin{document}
\title{Compact Graph Structure Learning via Mutual Information Compression}


\author{Nian Liu}
\email{nianliu@bupt.edu.cn}
\affiliation{%
  \institution{Beijing University of Posts and Telecommunications}
  \country{China}}

\author{Xiao Wang}
\email{xiaowang@bupt.edu.cn}
\affiliation{%
  \institution{Beijing University of Posts and Telecommunications}
  \institution{Peng Cheng Laboratory}
  \country{China}
  }

\author{Lingfei Wu}
\email{lwu@email.wm.edu}
\affiliation{%
  \institution{JD.COM Silicon Valley Research Center}
  \country{United States}}

\author{Yu Chen}
\email{hugochen@fb.com}
\affiliation{%
  \institution{Facebook AI}
  \country{United States}}

\author{Xiaojie Guo}
\email{xguo7@gmu.edu}
\affiliation{%
  \institution{JD.COM Silicon Valley Research Center}
  \country{United States}}

\author{Chuan Shi}
\authornote{Corresponding author.}
\email{shichuan@bupt.edu.cn}
\affiliation{%
  \institution{Beijing University of Posts and Telecommunications}
  \institution{Peng Cheng Laboratory}
  \country{China}
  }







\begin{abstract}
Graph Structure Learning (GSL) recently has attracted considerable attentions in its capacity of optimizing graph structure as well as learning suitable parameters of Graph Neural Networks (GNNs) simultaneously. Current GSL methods mainly learn an optimal graph structure (final view) from single or multiple information sources (basic views), however the theoretical guidance on what is the optimal graph structure is still unexplored. In essence, an optimal graph structure should only contain the information about tasks while compress redundant noise as much as possible, which is defined as "minimal sufficient structure", so as to maintain the accurancy and robustness. How to obtain such structure in a principled way? In this paper, we theoretically prove that if we optimize basic views and final view based on mutual information, and keep their performance on labels simultaneously, the final view will be a minimal sufficient structure. With this guidance, we propose a \textbf{\textit{Co}}mpact \textbf{\textit{GSL}} architecture by MI compression, named \textbf{\textit{CoGSL}}. Specifically, two basic views are extracted from original graph as two inputs of the model, which are refinedly reestimated by a view estimator. Then, we propose an adaptive technique to fuse estimated views into the final view. Furthermore, we maintain the performance of estimated views and the final view and reduce the mutual information of every two views. To comprehensively evaluate the performance of CoGSL, we conduct extensive experiments on several datasets under clean and attacked conditions, which demonstrate the effectiveness and robustness of CoGSL.
\end{abstract}



\maketitle

\section{Introduction}
Graph is capable of modeling real systems in diverse domains varying from natural language and images to network analysis. Nowadays, as an emerging technique, Graph Neural Networks (GNNs) \cite{gcn, gat, graphsage} have achieved great success with their characteristic message passing scheme \cite{messagepassing} that aims to aggregate information from neighbors continually. So far, GNNs have shown superior performance in a wide range of applications, such as node classification \cite{jump, gin} and link prediction \cite{link1, link2}.

It is well known that the performance of GNNs is closely related to the quality of given graphs \cite{lds}. However, due to the complexity of real information sources, the quality of graphs is often unreliable \cite{marsden1990network}. On one hand, we are not always provided with graph structures, such as in natural language processing \cite{nlp1, nlp2} or computer vision \cite{cv1, cv2}. In these cases, graphs are constructed by involving prior knowledge, which is sometimes error-prone. On the other hand, even though interactions between objects are extracted, spurious edges are usually inevitably existed in graphs. For example, it is very hard to analyze the molecular structure of unknown proteins \cite{alphafold}, so they are prone to be modeled with wrong or useless connections. Furthermore, graphs sometimes suffer from malicious attacks, such that original structures are fatally destroyed. With attacked graphs, GNNs will be very vulnerable. As a result, various drawbacks are prevalent in real graphs, which prohibits original structure from being the optimal one for downstream tasks.

Recently, graph structure learning (GSL) has aroused considerable attentions, which aims to learn optimal graph structure and parameters of GNNs simultaneously \cite{gsl_survey}. Current GSL methods can be roughly divided into two categories, single-view \cite{lds, neuralsparse, glcn, prognn} based and multi-view based \cite{gen, geom, idgl}. For the former, they usually estimate the optimal structure from one view, i.e., the given adjacency matrix, by forcing the learned structure to accord with some properties. For instance, Pro-GNN \cite{prognn} learns the graph structure with low rank, sparsity and feature smoothness constraints. For the later, considering that the measurement of an edge based on only one view may be biased, they aim to extract multiple basic views from original structure, and then comprehensively estimate the final optimal graph structure based on these views. As an example, IDGL \cite{idgl} constructs the structure by two type of views: normalized adjacency matrix and similarity matrix calculated with node embeddings. Multi-view based methods are able to utilize multifaceted knowledge to make the final decision on GSL.

Here, we focus on in-depth analysis of multi-view based GSL methods, and aim to answer one fundamental question: \textit{how can we estimate the optimal graph structure from multiple views in a principled way?} Despite that multiple views based GSL methods are considered as a  promising solution, there is still a lack of theoretical guidance to determine what is "optimal" in principle. In essence, an optimal graph structure should only contain the most concise information about downstream tasks (e.g., node labels), no more and no less, so that it can conduct the most precise prediction on labels. If the learned structure absorbs the information of labels as well as additional irrelevance from basic views, this structure is more prone to  adversarial attacks when small perturbations are deployed on these irrelevant parts. While if the learned structure only holds limited information about labels, the model probably fails to support downstream tasks. In summary, the optimal structure should contain minimal but sufficient information of labels, and we call it \textit{minimal sufficient structure}, which makes a well balance between effectiveness and robustness. Other GSL methods mainly focus on the performance, while neglecting the compactness of structure. Hence, the structures learnt from them inevitably contain redundant noise, and are also vulnerable to perturbations.


However, it is technically challenging to obtain a minimal sufficient graph structure. Particularly, two obstacles need to be addressed. (1) How to ensure the minimum and sufficiency of the final view? To achieve the sufficiency, the final view should be fully guided by labels, which makes it contain the information about labels as much as possible. And for the minimum, considering that the final view extracts information from basic views, we need to constrain the information flow from basic views to final view, which avoids irrelevant information and contributes to the conciseness of the final view. Therefore, to be minimal and sufficient, we we need to rethink on how to formulate the relations among basic views, final view and labels. (2) How to ensure the effectiveness of basic views? Considering that basic views are the information source of final view, it is vital to guarantee the quality of basic views. On one hand, basic views are also needed to contain the information about labels, which can fundamentally guarantee the performance of final view. On the other hand, these views also should be independent of each other, so that they can eliminate the redundancy and provide diverse knowledge about labels for final view. However, it is hard to guarantee the raw basic views satisfy these requirements, implying that we need to reestimate them..

In this paper, we study the problem of GSL with information theory, and propose CoGSL, a framework to learn compact graph structure with mutual information compression. Specifically, we first carefully extract two basic views from original structure as inputs, and design a view estimator to properly adjust basic views. With the estimated basic views, we propose a novel adaptive non-parameter fusion mechanism to get the final view. In this mechanism, the model will assign weights to basic views according to its predictions on nodes. If it gives a more confident prediction on one view, this view will be assigned with a larger weight. Then, we propose a formal definition \textit{minimal sufficient structure}. And we theoretically prove that if the performances of basic views and final view are guaranteed, we need to minimize the mutual information (MI) between every two views simultaneously. To effectively evaluate the MI between different views, we deploy a MI estimator implemented based on InfoNCE loss \cite{cpc}. In the end, we adopt a three-fold optimization to practically initialize the principles. Our contributions are summarized as follows:
\begin{itemize}[leftmargin=*]
    \item To our best knowledge, we are the first to utilize information theory to study the optimal structure in GSL. We propose the concept of "minimal sufficient structure", which aims to learn the most compact structure relevant to downstream tasks in principle, no more and no less, so as to provide a better balance between accuracy and robustness.
    \item We theoretically prove that the minimal sufficient graph structure heavily depends on modeling the relationships among different views and labels. Based on this, we propose CoGSL, a novel framework to learn compact graph structure via mutual information compression.
    \item We validate the effectiveness of CoGSL compared with state-of-the-art methods on seven datasets. Additionally, CoGSL also outperforms other GSL methods on attacked datasets, which further demonstrates the robustness of CoGSL.
\end{itemize}
\section{related work}
\textbf{Graph Neural Network}. Graph neural networks (GNNs) have attracted considerable attentions recently, which can be broadly divided into two categories, spectral-based and spatial-based. Spectral-based GNNs are inheritance of graph signal processing, and define graph convolution operation in spectral domain. For example, \cite{spe1} utilizes Fourier bias to decompose graph signals; \cite{spe2} employs the Chebyshev expansion of the graph Laplacian to improve the efficiency. For another line, spatial-based GNNs greatly simplify above convolution by only focusing on neighbors. For example, GCN \cite{gcn} simply averages information of one-hop neighbors. GraphSAGE \cite{graphsage} only randomly fuses a part of neighbors with various poolings. GAT \cite{gat} assigns different weights to different neighbors. More detailed surveys can be found in \cite{wu2021comprehensive}.

\textbf{Graph Structure Learning}. Graph structure learning aims to estimate a better structure for original graph, which can date back to previous works in network science \cite{handcock2010modeling, lusher2013exponential}. In this paper, we mainly focus on GNN based graph structure learning models. LDS \cite{lds} jointly optimizes the probability for each node pair and GNN in a bilevel way. Pro-GNN \cite{prognn} aims to obtain a clear graph by deploying some regularizers, such as low-rank, sparsity and feature smoothness. IDGL \cite{idgl} casts the GSL as a similarity metric learning problem. GEN \cite{gen} presents an iterative framework based on Bayesian inference. However, these methods do not provide a theoretical view to show what the optimal structure is.
\section{the proposed model}
\begin{figure*}[t]
  \centering
  \includegraphics[scale=0.3]{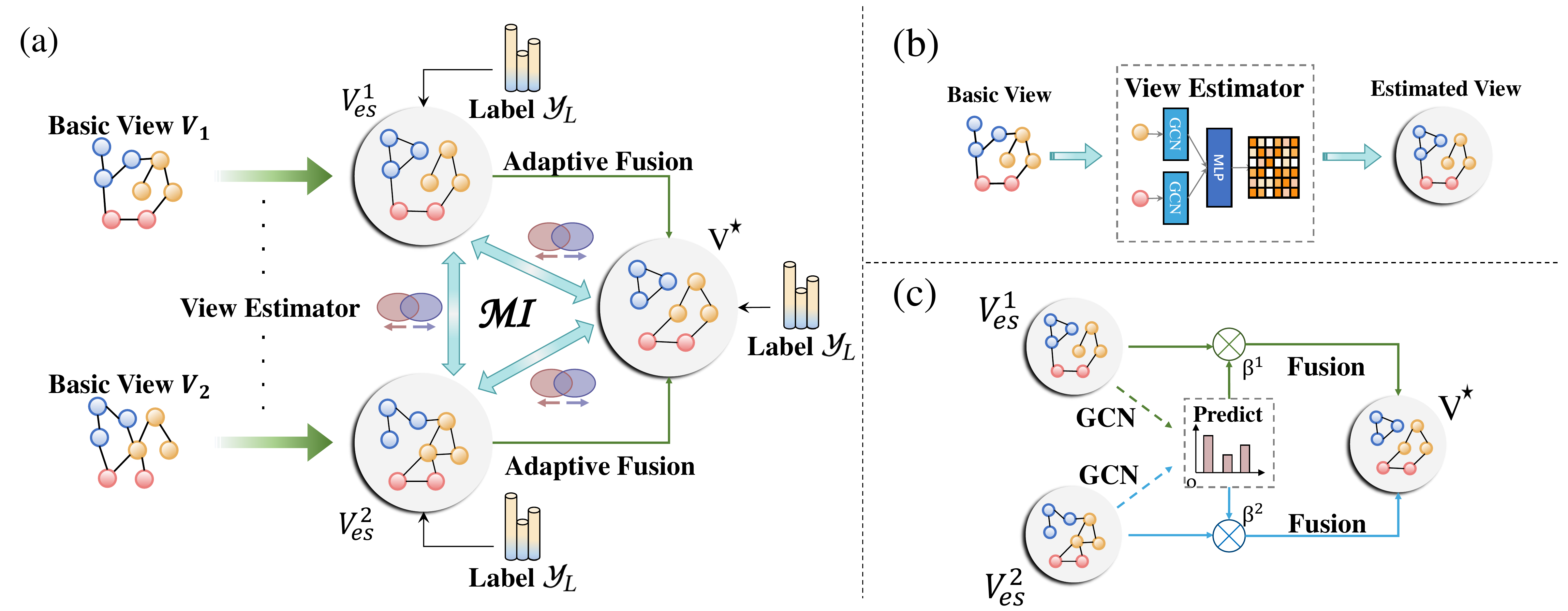}
  \caption{The overview of our proposed CoGSL. (a) Model framework. (b) View estimator. (c) Adaptive fusion.}
  \Description{}
  \label{model}
\end{figure*}

In this section, we elaborate the proposed model CoGSL for GSL, and the overall architecture is shown in Fig.~\ref{model}(a). Our model begins with two basic views. Then, we design a view estimator to optimize two basic views separately. With two estimated views, we propose an adaptive fusion technique to generate final view based on the confidence of predictions. Next, we formally propose the concept "minimal sufficient structure", and make a proposition to guarantee the final view to be minimal and sufficient.

\subsection{Problem definition}
Let $\mathcal{G=(\mathcal{V}, \xi)}$ represent a graph, where $\mathcal{V}$ is the set of N nodes and $\xi$ is the set of edges. All edges formulate an original adjacency matrix $\bm{A}\in\mathbb{R}^{N\times N}$, where $A_{ij}$ denotes the relation between nodes $v_i$ and $v_j$. Graph $\mathcal{G}$ is often assigned with node feature matrix $\bm{X}=[x_1, x_2,\dots, x_N]\in\mathbb{R}^{N\times D}$, where $x_i$ means the D dimensional feature vector of node $i$. In semi-supervised classification, we only have a small part of nodes with labels $\mathcal{Y}_L$. The traditional goal of graph structure learning for GNNs is to simultaneously learn an optimal structure and GNN parameters to boost downstream tasks. 

As one typical architecture, GCN \cite{gcn} is usually chosen as the backbone, which iteratively aggregates neighbors' information. Formally, the $k^{th}$ GCN layer can be written as:
\begin{equation}
    GCN(\bm{A}, \bm{H}^{(k)}) = \bm{D}^{-1/2}\bm{AD}^{-1/2}\bm{H}^{(k-1)}\bm{W}^{k},
\end{equation}
where \bm{$D$} is the degree matrix of \bm{$A$}, and $\bm{W}^{k}$ is weight matrix. $\bm{H}^{(k)}$ represents node embeddings in the $k^{th}$ layer, and $\bm{H}^{(0)}=\bm{X}$. In this paper, we simply utilize $GCN(\bm{V}, \bm{H})$ to represent this formula, where $\bm{V}$ is some view and $\bm{H}$ is the node features or embeddings.

\subsection{The selection of basic views}
Given a graph G, CoGSL starts from extracting different structures. In this paper, we mainly investigate four widely-studied structures: (1) Adjacency matrix, which reflects the local structure; (2) Diffusion matrix, which represents the stationary transition probability from one node to other nodes and provides a global view of graph. Here, we choose Personal PageRank (PPR), whose closed-form solution \cite{mvgrl} is $    \bm{S}=\alpha(\bm{I}-(1-\alpha)\bm{D}^{-1/2}\bm{AD}^{-1/2})^{-1}$, where $\alpha\in(0, 1]$ denotes teleport probability in a random walk, \bm{$I$} is a identity matrix, and \bm{$D$} is the degree matrix of \bm{$A$}; (3) Subgraph, which is special for large graph. We randomly keep a certain number of edges to generate a subgraph; (4) KNN graph, which reflects the similarity in feature space. We utilize original features to calculate cosine similarity between each node pair, and retain top-k similar nodes for each node to construct KNN graph.

These four views contain the different properties from various angles, and we carefully select two of them as two basic views \bm{${V_1}$} and \bm{${V_2}$}, which are the inputs of CoGSL. 

\subsection{View Estimator}
\label{Estimator}
Given two basic views \bm{$V_1$} and \bm{$V_2$}, we need to further polish them so that they are more flexible to generate the final view. Here, we devise a view estimator for each basic view, shown in Fig.~\ref{model}(b). Specifically, for basic view \bm{$V_1$}, we first conduct a GCN \cite{gcn} layer to get embeddings $\bm{Z^1}\in\mathbb{R}^{N\times d_{es}}$:
\begin{equation}
\label{view_1}
    \bm{Z^1} = \sigma(GCN(\bm{V_1}, \bm{X})),
\end{equation}
where $\sigma$ is non-linear activation. With embedding \bm{$Z^1$}, probability of an edge between each node pair in \bm{$V_1$} can be reappraised. For target node $i$, we concatenate its embedding \bm{$z^1_i$} with embedding \bm{$z^1_j$} of another node $j$, which is followed by a MLP layer:
\begin{equation}
    w^1_{ij} = \bm{W_1}\cdot[\bm{z^1_i}||\bm{z^1_j}]+b_1,
\end{equation}
where $w^1_{ij}$ denotes the weight between $i$ and $j$, $\bm{W_1}\in\mathbb{R}^{2d_{es}\times1}$ is mapping vector, and $b_1\in\mathbb{R}^{2d_{es}\times1}$ is the bias vector. Then, we normalize the weights for node $i$ to get the probability $p^1_{ij}$ between node $i$ and other node $j$. Moreover, to alleviate space and time expenditure, we only estimate limited scope $S^1$. For example, for adjacency matrix, KNN or subgraph, we only inspect their k-hop neighbors, and for diffusion matirx, we only reestimate top-h neighbors for each node according to PPR values. Here, h and k are hyper-parameters. So, $p^1_{ij}$ is calculated as:
\begin{equation}
    p^1_{ij} = \frac{\exp(w^1_{ij})}{\sum_{k\in S^1} \exp(w^1_{ik})}.
    \label{prob_new}
\end{equation}
In this way, we construct a probability matrix $\bm{P^1}$, where each entry is calculated by eq.~\eqref{prob_new}. Combined with original structure, the estimated view is as follows: 
\begin{equation}
\label{view_2}
    \bm{V^1_{es}}=\bm{V_1}+\mu^1\cdot \bm{P^1},
\end{equation}
where $\mu^1\in(0, 1)$ is a combination coefficient, and the $i$th row of \bm{$V^1_{es}$}, denoted as \bm{$V^1_{es\_i}$}, shows new neighbors of node $i$ in the estimated view. Estimating \bm{$V_2$} is similar to \bm{$V_1$} but with a different set of parameters, and we can get the estimated view \bm{$V^2_{es}$} finally.

\subsection{View Fusion}
\label{View Fusion}
Then, the question we would like to answer is: given two estimated views, how can we effectively fuse them in an adaptive way for each node? We utilize the confidence of predictions as the evidence to fuse estimated views, and assign a larger weight to the more confident view. In this way, the final view can make a more confident prediction and get more effectively trained. Specifically, we first utilize two-layer GCNs to obtain predictions of each view:
\begin{equation}
\begin{aligned}
\label{prediction}
    \bm{O^1}&=softmax(GCN(\bm{V^1_{es}},\ \sigma(GCN(\bm{V^1_{es}}, \bm{X})))),\\
    \bm{O^2}&=softmax(GCN(\bm{V^2_{es}},\ \sigma(GCN(\bm{V^2_{es}}, \bm{X})))),
\end{aligned}
\end{equation}
where $\sigma$ is activation function, and for node $i$, its predictions on these two views are \bm{$o^1_i$} and \bm{$o^2_i$}. Next, we show two cases to analyze how to assign weights to estimated views based on node predictions. In the first case, \bm{$o^1_i$} presents a sharp distribution (e.g. [0.8, 0.1, 0.1] for three-way classification), while \bm{$o^2_i$} is a smoother distribution (e.g. [0.4, 0.3, 0.3]). During the fusion, if we assign larger weight to \bm{$V^2_{es\_i}$}, the final view still give an uncertain result, and the model cannot be trained effectively. So in this case, we suggest to emphasize \bm{$V^1_{es\_i}$}. For the second case, predictions \bm{$o^1_j$} and \bm{$o^2_j$} of node $j$ are [0.5, 0.4, 0.1] and [0.5, 0.25, 0.25], respectively. Although they have the same maximal value, there is a larger margin between maximal value and submaximal value in \bm{$o^2_j$}, so \bm{$V^2_{es\_j}$} is a more confident view. In conclusion, if one estimated view has a higher maximal value and a larger margin between maximal value and submaximal value, it is a more confident view, which should be assigned with larger weight. With above analysis, we propose an adaptive fusion for each node, shown in Fig.~\ref{model}(c). Specifically, we focus on node $i$ to explain our fusion mechanism. First, we calculate the importance $\pi^1$ of \bm{$V^1_{es\_i}$}:
\begin{equation}
    \label{ff}
    \pi_i^1 = e^{\epsilon\left(\lambda\log o^1_{i, m}+(1-\lambda)\log(o^1_{i, m}-o^1_{i, sm})\right)},
\end{equation}
where $o^1_{i, m}$ and $o^1_{i, sm}$ denote the maximal and submaximal values of prediction \bm{$o^1_i$}, $\epsilon$ and $\lambda$ are hyper-parameters. The eq.~\eqref{ff} has three advantages: (1) If the prediction of one view has a higher maximal value and a larger margin between maximal and submaximal values, this view is prone to make confident decision, and it will lead the fusion. (2) This mechanism fully takes account of each node, so it achieves the adaptive fusion. (3) This mechanism of calculating importance does not introduce new parameters, so it alleviates over-fitting to some extent. Similarly, we can get the importance $\pi_i^2$ of \bm{$V^2_{es\_i}$}. Next, we normalize the importance and get the weights:
\begin{equation}
    \beta_i^1 = \frac{\pi_i^1}{\pi_i^1+\pi_i^2}\quad and\quad \beta_i^2 = \frac{\pi_i^2}{\pi_i^1+\pi_i^2}.
\end{equation}
Finally, we generate the final view for node $i$ based on weights:
\begin{equation}
\label{end}
    \bm{V^{\star}_i} = \beta_i^1\cdot \bm{V^1_{es\_i}} + \beta_i^2\cdot \bm{V^2_{es\_i}}.
\end{equation}
We likewise copy above operations to get the fusion for other nodes, and the final view \bm{$V^{\star}$} is the combination of these fusion results. 


\subsection{Learning a minimal sufficient structure $V^{\star}$}
\subsubsection{Theoretical Motivation and Formulation}
Up to now, we have discussed how to adaptively fuse basic views to generate final view \bm{$V^{\star}$}, which will be used for downstream tasks. The next issue is \textit{how to guide the training of\quad\bm{$V^{\star}$}, and what the principle should be obeyed to deal with the relations between basic views and final view?} Please review that we expect the learnt \bm{$V^{\star}$} only contains the message about labels and filters superfluous noise. In other words, we seek a \bm{$V^{\star}$} that is the minimal sufficient statistics \cite{minimal} to label $\mathcal{Y}_L$ in information theory. The formal definition is given as follows:
\begin{myDef}
    (Minimal Sufficient Structure) Given two variables $U$ and $V$, I(U; V) means mutual information (MI), H(U) is entropy, and H(U| V) is conditional entropy. A structure \bm{$V^{\star}$} is the minimal sufficient structure if and only if I(\bm{$V^{\star}$}; $\mathcal{Y}_L$) = H($\mathcal{Y}_L$) and H(\bm{$V^{\star}$}| $\mathcal{Y}_L$) = 0.
\end{myDef}

In this definition, $I(\bm{V^{\star}}; \mathcal{Y}_L) = H(\mathcal{Y}_L)$ means $\bm{V^{\star}}$ shares all the information about $H(\mathcal{Y}_L)$, and $H(\bm{V^{\star}}| \mathcal{Y}_L) = 0$ guarantees that $\bm{V^{\star}}$ does not contain any other information except $H(\mathcal{Y}_L)$. To gain such minimal sufficient structure, we have the following proposition.
\begin{Proposition}
\label{prop}
    Given the estimated basic views \bm{$V^1_{es}$} and \bm{$V^2_{es}$}, final view \bm{$V^{\star}$}, and labels $\mathcal{Y}_L$ related to downstream task, \bm{$V^{\star}$} is a minimal sufficient structure to $\mathcal{Y}_L$ if the following two principles are satisfied: \\
    1.\indent $I(\bm{V^1_{es}}; \mathcal{Y}_L)=I(\bm{V^2_{es}}; \mathcal{Y}_L)=I(\bm{V^{\star}}; \mathcal{Y}_L)=H(\mathcal{Y}_L)$ \\
    2.\indent minimize\ \ $I(\bm{V^1_{es}}; \bm{V^2_{es}})+I(\bm{V^1_{es}}; \bm{V^{\star}})+I(\bm{V^2_{es}};\bm{V^{\star}})$
\end{Proposition}

For node classification task, the first principle will build the relationships between $\bm{V^1_{es}}$, $\bm{V^2_{es}}$, $\bm{V^{\star}}$ and $\mathcal{Y}_L$ based on MI. In this way, the information of $\mathcal{Y}_L$ will be totally contained in $\bm{V^1_{es}}$, $\bm{V^2_{es}}$ and $\bm{V^{\star}}$, which makes them hold sufficient information about $\mathcal{Y}_L$. Meanwhile, we perform the second principle to constrain the shared information among views, which finally realizes a minimal \bm{$V^{\star}$}. Now, we prove the effect of the second principle:

\begin{proof}
At the beginning, we introduce some basic properties in information theory \cite{property}, which describe entropy $H(X)$, conditional entropy $H(Y| X)$, joined entropy $H(X, Y)$, mutual information $I(X; Y)$ and conditional mutual information $I(X; Z|Y)$. \\
(1) Nonnegativity: \\
\centerline{$H(X| Y)\geq0; I(X; Y| Z)\geq0$} \\
(2) Chain rule of entropy and MI: \\
\centerline{$H(X, Y)=H(X)+H(Y|X)$} \\
\centerline{$I(X; Y, Z)=I(X; Y)+I(X; Z|Y)$} \\
(3) Multivariate mutual information: \\
\centerline{$I(X_1; X_2;\dots;X_{n+1})=I(X_1;\dots;X_n)-I(X_1;\dots;X_n|X_{n+1})$}

Then, we have the following proof:

First, we have $I(\bm{V^{\star}}; \bm{V^1_{es}}; \bm{V^2_{es}})>0$, because these three views share the information of $\mathcal{Y}_L$ at least, which is guaranteed by the first principle. So, we have:

\begin{align*}
\label{11}
&I(\bm{V^1_{es}}; \bm{V^2_{es}})+I(\bm{V^1_{es}}; \bm{V^{\star}})+I(\bm{V^2_{es}}; \bm{V^{\star}})\\
>\ &I(\bm{V^1_{es}}; \bm{V^2_{es}})+I(\bm{V^1_{es}}; \bm{V^{\star}})+I(\bm{V^2_{es}}; \bm{V^{\star}})-2I(\bm{V^{\star}}; \bm{V^1_{es}}; \bm{V^2_{es}}) \\
=\ &I(\bm{V^1_{es}}; \bm{V^2_{es}})+I(\bm{V^{\star}}; \bm{V^1_{es}}|\bm{V^2_{es}})+I(\bm{V^{\star}}; \bm{V^2_{es}}|\bm{V^1_{es}})\\
=\ &I(\bm{V^{\star}}; \bm{V^1_{es}}; \bm{V^2_{es}})+I(\bm{V^1_{es}}; \bm{V^2_{es}}|\bm{V^{\star}})+I(\bm{V^{\star}}; \bm{V^1_{es}}|\bm{V^2_{es}})\\
&+I(\bm{V^{\star}}; \bm{V^2_{es}}|\bm{V^1_{es}})\\
=\ &I(\bm{V^{\star}}; \bm{V^1_{es}})+I(\bm{V^{\star}}; \bm{V^2_{es}}|\bm{V^1_{es}})+I(\bm{V^1_{es}}; \bm{V^2_{es}}|\bm{V^{\star}})\\
=\ &I(\bm{V^{\star}}; \bm{V^1_{es}}, \bm{V^2_{es}})+I(\bm{V^1_{es}}; \bm{V^2_{es}}|\bm{V^{\star}})\\
=\ &H(\bm{V^{\star}})-H(\bm{V^{\star}}|\bm{V^1_{es}}\bm{V^2_{es}})+I(\bm{V^1_{es}}; \bm{V^2_{es}}|\bm{V^{\star}})
\end{align*}

In the last step, $H(\bm{V^{\star}}|\bm{V^1_{es}}\bm{V^2_{es}}) = 0$. This is because \bm{$V^{\star}$} is an adaptive combination of \bm{$V^1_{es}$} and \bm{$V^2_{es}$}, and if \bm{$V^1_{es}$} and \bm{$V^2_{es}$} are known, there is no uncertainty in \bm{$V^{\star}$}. Thus, we have:
\begin{equation}
\label{22}
    I(\bm{V^1_{es}}; \bm{V^2_{es}})+I(\bm{V^1_{es}}; \bm{V^{\star}})+I(\bm{V^2_{es}}; \bm{V^{\star}}) > H(\bm{V^{\star}})+I(\bm{V^1_{es}}; \bm{V^2_{es}}|\bm{V^{\star}}).
\end{equation}
Furthermore, we can expand $H(\bm{V^{\star}})$ to $H(\bm{V^{\star}}, \mathcal{Y}_L)$, because the information of $\mathcal{Y}_L$ is totally contained in $\bm{V^{\star}}$, according to the first principle. Next, we have the following derivation:
\begin{equation}
\begin{aligned}
\label{33}
&H(\bm{V^{\star}})+I(\bm{V^1_{es}}; \bm{V^2_{es}}|\bm{V^{\star}}) \\
=&H(\bm{V^{\star}}, \mathcal{Y}_L)+I(\bm{V^1_{es}}; \bm{V^2_{es}}|\bm{V^{\star}})\\
=&H(\mathcal{Y}_L)+H(\bm{V^{\star}}|\mathcal{Y}_L)+I(\bm{V^1_{es}}; \bm{V^2_{es}}|\bm{V^{\star}}).
\end{aligned}
\end{equation}
According to eq.~\eqref{22} and eq.~\eqref{33}, we have:
\begin{equation}
\begin{aligned}
\label{final}
&I(\bm{V^1_{es}}; \bm{V^2_{es}})+I(\bm{V^1_{es}}; \bm{V^{\star}})+I(\bm{V^2_{es}}; \bm{V^{\star}}) \\
>\ &H(\mathcal{Y}_L)+H(\bm{V^{\star}}|\mathcal{Y}_L)+I(\bm{V^1_{es}}; \bm{V^2_{es}}|\bm{V^{\star}}),
\end{aligned}
\end{equation}

In inequation~\ref{final}, $H(\bm{V^{\star}}|\mathcal{Y}_L)\geq 0$ and $I(\bm{V^1_{es}}; \bm{V^2_{es}}|\bm{V^{\star}})\geq 0$ according to nonnegativity shown above. $H(\mathcal{Y}_L)$ is a constant, because the information of $\mathcal{Y}_L$ is fixed. Ideally, both of $H(\bm{V^{\star}}|\mathcal{Y}_L)$ and $I(\bm{V^1_{es}}; \bm{V^2_{es}}|\bm{V^{\star}})$ equal to 0 by continuously minimizing the original formula. This means given labels $\mathcal{Y}_L$, $\bm{V^{\star}}$ does not include other information any more, and become a minimal sufficient structure. Meanwhile, $\bm{V^1_{es}}$ and $\bm{V^2_{es}}$ only share the information of $\bm{V^{\star}}$. So, $\bm{V^1_{es}}$ and $\bm{V^2_{es}}$ only share the message about $\mathcal{Y}_L$, and they will provide the most diverse knowledge for $V^{\star}$.
\end{proof}

\subsubsection{Iterative Optimization}
Based on Proposition~\ref{prop}, we design a three-fold optimization objective: (1) Optimize parameters $\Theta$ of classifiers for each view to improve the accuracy on $\mathcal{Y}_L$; (2) Optimize parameters $\Phi$ of MI estimator to approach the real MI value; (3) Optimize parameters $\Omega$ of view estimator to maintain classification accuracy and minimize the MI between every two views simultaneously. 

\textbf{Optimize $\Theta$.} Please recall that predictions of \bm{$V^1_{es}$} and \bm{$V^2_{es}$} have been obtained according to eq.~\eqref{prediction}, denoted as \bm{$O^1$} and \bm{$O^2$}. Similarly, we also can get the predictions of \bm{$V^{\star}$}:
\begin{equation}
    \bm{O^{\star}}=softmax(GCN(\bm{V^{\star}},\ \sigma(GCN(\bm{V^{\star}}, \bm{X})))).
    \label{prediction1}
\end{equation}
The parameters of $GCN$s involved in eq~\eqref{prediction} and eq.~\eqref{prediction1} are regarded as the parameters $\Theta$ of classifiers together. $\Theta$ can be optimized by evaluating the cross-entropy error over $\mathcal{Y}_L$
\begin{equation}
\label{trian_theta}
    \min\limits_{\Theta} \mathcal{L}_{cls} = -\sum\limits_{\bm{O}\in\{\bm{O^1}, \bm{O^2}, \bm{O^{\star}}\}}\sum\limits_{v_i\in{\mathcal{Y}_L}}y_i\ln{\bm{o_i}},
\end{equation}
where $y_i$ is the label of node $v_i$, and \bm{$o_i$} is its prediction.

\begin{figure}[t]
  \centering
  \includegraphics[scale=0.6]{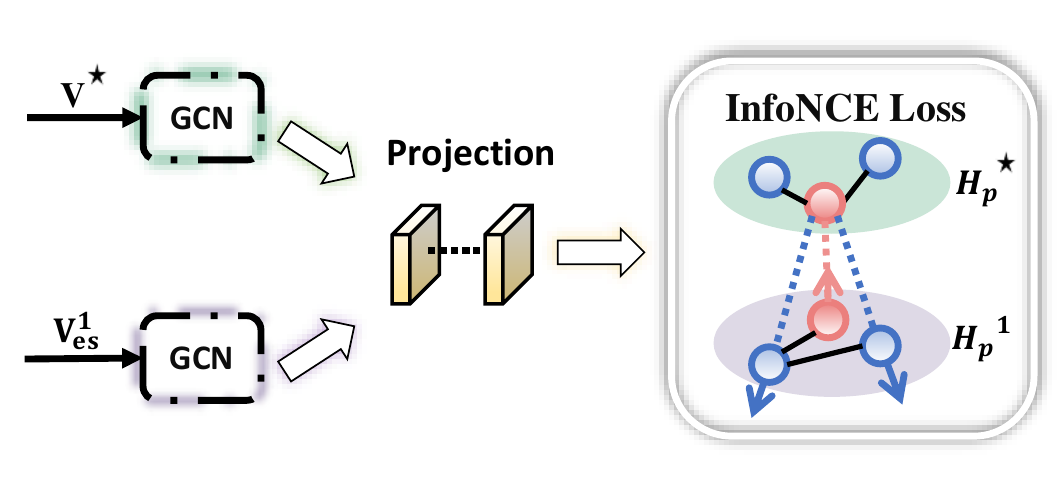}
  \caption{An illustration to show the process of MI estimator. (Take $\bm{V^{\star}}$ and $\bm{V^1_{es}}$ for example)}
  \Description{}
  \label{infonce}
\end{figure}

\textbf{Optimize $\Phi$.} In view of the second principle, we need to minimize MI values of every two views. However, estimating precise MI is hard \cite{hardcomp}. Recently, InfoNCE \cite{cpc,simclr,moco} has been proved as a lower bound of real MI. If InfoNCE loss is minimized, we can approximately approach the real MI. Here, we design a corresponding MI estimator. The whole process of MI estimator is shown in Fig. \ref{infonce}. Specifically, for \bm{$V^{\star}$}, we first conduct one-layer GCN to get node embeddings based on \bm{$V^{\star}$}:
\begin{equation}
\begin{aligned}
\label{mi_conv}
    \bm{H^{\star}}&=\sigma(GCN(\bm{V^{\star}}, \bm{X})), 
\end{aligned}
\end{equation}
where $\sigma$ is PReLU activation, $X$ is the feature matrix.The embeddings $\bm{H^{1}}$ and $\bm{H^{2}}$ based on \bm{$V^1_{es}$} and \bm{$V^2_{es}$} can be obtained in a similar way. The parameters of above three GCNs are different, but \{$\bm{H^{\star}}, \bm{H^1}, \bm{H^2}$\} have the same embedding dimension. Then we deploy a shared two-layer MLP to project these embeddings into the same space where MI estimation is employed, and get the projected embeddings \bm{$H^{\star}_{p}$}, \bm{$H^1_{p}$} and \bm{$H^2_{p}$}, respectively. For example, the projected embeddings \bm{$H^{\star}_{p}$} is got as follows:
\begin{equation}
\begin{aligned}
    \bm{H^{\star}_{p}}&=\bm{W^1}\cdot \sigma(\bm{W^0}\cdot \bm{H^{\star}}+b^0)+b^1, \\
\end{aligned}
\end{equation}
where $\sigma$ is non-linear activation, and $\{\bm{W^0}, \bm{W^1}, b^0, b^1\}$ are shared parameters. Then, inspired by GCA \cite{gca}, we take \bm{$H^{\star}_{p}$} and \bm{$H^1_{p}$} for example to give the InfoNCE loss as follows:

\begin{equation}
\begin{aligned}
\label{contra_loss}
    &L(\bm{V^{\star}}, \bm{V^1_{es}}) \\
    =&-\frac{1}{2|B|}\sum\limits_{i=1}^{|B|}\left[\log\frac{e^{sim(\bm{h^{\star}_{p_i}}, \bm{h^1_{p_i})}/\tau}}{e^{sim(\bm{h^{\star}_{p_i}}, \bm{h^1_{p_i}})/\tau}+\sum_{k\neq i}e^{sim(\bm{h^{\star}_{p_i}},\bm{ h^1_{p_k}})/\tau}} \right.\\
    &\left.+\log\frac{e^{sim(\bm{h^1_{p_i}}, \bm{h^{\star}_{p_i}})/\tau}}{e^{sim(\bm{h^1_{p_i}}, \bm{h^{\star}_{p_i}})/\tau}+\sum_{j\neq i}e^{sim(\bm{h^1_{p_i}}, \bm{h^{\star}_{p_j}})/\tau}}\right],
\end{aligned}
\end{equation}
where $sim(u, v)$ is cosine similarity of vector $u$ and $v$, and $\tau$ is temperature coefficient. \bm{$h^{\star}_{p_i}$} and \bm{$h^1_{p_i}$} are the projected embeddings of node $i$ based on \bm{$V^1_{es}$} and \bm{$V^2_{es}$}, respectively. $B$ is a batch of nodes that randomly sampled. This formula means if we maximize the similarity of embeddings of the same node but from different views, while minimize the similarity with other nodes in the same batch, we can approximatively approach real MI between \bm{$V^{\star}$} and \bm{$V^1_{es}$}. Similarly, we can calculate $L(\bm{V^{\star}}, \bm{V^2_{es}})$ and $L(\bm{V^1_{es}, V^2_{es}})$, and the objective to optimize MI estimator is shown here:
\begin{equation}
\label{mi_loss}
    \mathcal{L}_{MI}=L(\bm{V^{\star}, V^1_{es}})+L(\bm{V^{\star}, V^2_{es}})+L(\bm{V^1_{es}, V^2_{es}}).
\end{equation}
By minimizing the above equation, the MI estimator $\Phi$, including parameters in eq.~\eqref{mi_conv} and the following shared MLP, is well trained.

\textbf{Optimize $\Omega$.} Given trained classifiers and MI estimator, we continuously optimize parameters $\Omega$ of view estimator. Under the guidance of proposition~\ref{prop}, we have the following loss:
\begin{equation}
\label{trian_omega}
    \min\limits_{\Omega} \mathcal{L}_{cls}-\eta\cdot\mathcal{L}_{MI},
\end{equation}
where $\eta$ is a balance parameter. With this optimization, \bm{$V^1_{es}$} and \bm{$V^2_{es}$} only share the information of \bm{$V^{\star}$}, and \bm{$V^{\star}$} only reserves useful information while filters noise as far as possible.

To effectively train the CoGSL, we alternatively and iteratively perform the above three-fold optimization, where a profile of the whole process is shown in appendix~\ref{alg}. We can optimize the proposed CoGSL via back propagation with stochastic gradient descent.

\begin{table*}[h]
  \caption{Quantitative results (\%$\pm\sigma$) on node classification.(bold: best; underline: runner-up)}
  \label{fenlei}
  \resizebox{0.9\textwidth}{!}{
  \begin{tabular}{c|c|cc|ccc|cccc|c}
    \bottomrule
    Datasets & Metric & DGI & GCA & GCN & GAT & GraphSAGE & LDS & Pro-GNN & IDGL & GEN & \textbf{CoGSL}\\
    \bottomrule
    \multirow{3}{*}{Wine}&
    F1-macro&93.6±0.8&94.5±2.7&94.1±0.6&93.6±0.4&96.3±0.8&93.4±1.0&\underline{97.3±0.3}&96.3±1.1&96.4±1.0&\textbf{97.9±0.3}\\
    &F1-micro&93.6±0.8&94.6±2.4&93.9±0.6&93.7±0.3&96.2±0.8&93.4±0.9&\underline{97.2±0.3}&96.2±1.1&96.3±1.0&\textbf{97.8±0.3}\\
    &AUC&99.5±0.1&97.8±1.4&99.6±0.2&97.8±0.2&99.4±0.4&99.0±0.1&99.5±0.1&\underline{99.6±0.1}&99.3±0.2&\textbf{99.7±0.1}\\
    \hline
    \multirow{3}{*}{Cancer}&
    F1-macro&85.7±1.9&93.4±1.2&93.0±0.6&92.2±0.2&92.0±0.5&83.1±1.5&93.3±0.5&93.1±0.9&\underline{94.1±0.8}&\textbf{94.6±0.3}\\
    &F1-micro&87.6±1.4&93.8±1.2&93.3±0.5&92.9±0.1&92.5±0.5&84.8±0.8&93.8±0.5&93.6±0.9&\underline{94.3±1.0}&\textbf{95.0±0.3}\\
    &AUC&95.2±2.4&97.9±0.6&\textbf{98.9±0.1}&96.9±0.3&96.9±0.5&90.6±0.9&97.8±0.2&98.1±0.3&98.3±0.3&\underline{98.5±0.1}\\
    \hline
    \multirow{3}{*}{Digits}&
    F1-macro&88.9±0.8&89.5±1.4&89.0±1.3&89.9±0.2&87.5±0.2&79.7±1.0&89.7±0.3&\underline{92.5±0.5}&91.3±1.3&\textbf{93.3±0.3}\\
    &F1-micro&89.0±0.8&89.6±1.5&89.1±1.3&90.0±0.2&87.7±0.2&80.2±0.9&89.8±0.3&\underline{92.6±0.5}&91.4±1.2&\textbf{93.3±0.3}\\
    &AUC&99.0±0.1&97.6±0.3&98.9±0.2&98.3±0.4&98.7±0.1&95.1±0.1&98.1±0.2&\underline{99.4±0.1}&98.4±0.9&\textbf{99.6±0.0}\\
    \hline
    \multirow{3}{*}{Polblogs}
    &F1-macro&90.9±0.4&95.0±0.2&95.1±0.4&94.1±0.1&93.3±2.5&94.9±0.3&94.6±0.6&94.6±0.7&\underline{95.2±0.6}&\textbf{95.5±0.1}\\
    &F1-micro&90.9±0.4&95.0±0.2&95.1±0.4&94.1±0.1&93.4±2.5&94.9±0.3&94.6±0.6&94.6±0.7&\underline{95.2±0.6}&\textbf{95.5±0.1}\\
    &AUC&96.4±0.3&98.2±0.2&\textbf{98.5±0.0}&97.4±0.1&98.1±0.1&98.1±0.4&98.3±0.2&98.2±0.2&98.0±0.6&\underline{98.3±0.1}\\
    \hline
    \multirow{3}{*}{Citeseer}
    &F1-macro&68.1±0.6&60.9±0.9&67.4±0.3&68.4±0.2&67.1±0.8&\underline{69.4±0.7}&63.1±0.7&69.2±0.9&68.7±0.5&\textbf{70.2±0.6}\\
    &F1-micro&72.1±0.6&64.5±1.1&70.1±0.2&72.2±0.2&70.1±0.7&72.2±0.7&65.6±0.8&\underline{72.6±0.6}&72.5±0.8&\textbf{73.4±0.8}\\
    &AUC&90.8±0.1&88.5±0.7&89.9±0.2&90.2±0.1&90.5±0.3&\underline{91.3±0.3}&88.2±0.3&91.1±0.4&88.4±0.5&\textbf{91.4±0.5}\\
    \hline
    \multirow{3}{*}{Wiki-CS}
    &F1-macro&56.4±0.1&67.1±1.3&68.8±1.7&\underline{70.1±0.1}&69.2±0.9&54.6±0.5&63.8±2.0
    &69.1±1.1&68.4±0.3&\textbf{72.3±0.6}\\
    &F1-micro&61.2±0.2&71.3±1.3&70.8±1.8&\underline{73.8±0.3}&72.2±0.7&53.7±0.5&68.3±1.2
    &72.7±0.8&71.1±0.9&\textbf{75.0±0.3}\\
    &AUC&91.8±0.1&93.2±0.4&95.2±0.3&\underline{95.6±0.1}&95.0±0.3&88.8±2.1&93.3±0.3&92.0±0.2&91.6±1.2&\textbf{96.4±0.2}\\
    \hline
    \multirow{3}{*}{MS Academic}
    &F1-macro&88.6±0.2&87.0±1.6&89.4±0.6&86.7±0.6&88.9±0.4&-&-&89.6±0.6&\underline{89.8±0.8}&\textbf{90.5±0.4}\\
    &F1-micro&91.4±0.2&89.8±1.2&91.9±0.5&89.0±0.4&91.1±0.2&-&-&91.9±0.5&\underline{92.0±0.5}&\textbf{92.4±0.5}\\
    &AUC&99.1±0.1&99.3±0.2&99.4±0.1&99.2±0.1&99.4±0.0&-&-&\textbf{99.6±0.1}&98.8±0.3&\underline{99.4±0.1}\\
    \hline
  \end{tabular}}
\end{table*}
\section{experiments}
\subsection{Experimental Setup}

\begin{figure*}[tbp]
\centering
\subfigure[Cancer]{
\label{dele_cancer}
\includegraphics[scale=0.25]{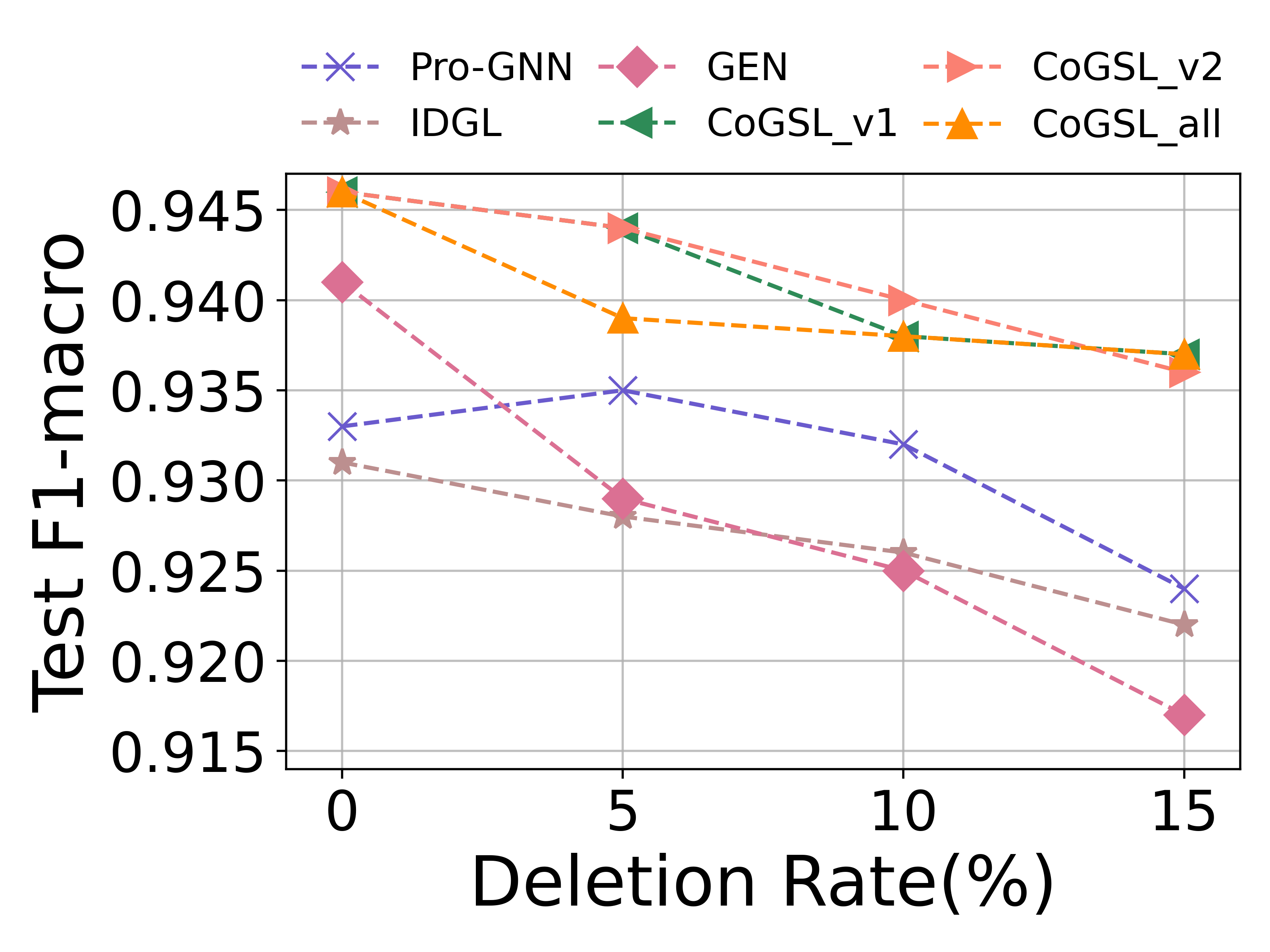}
}
\quad\quad
\subfigure[Citeseer]{
\label{dele_citeseer}
\includegraphics[scale=0.25]{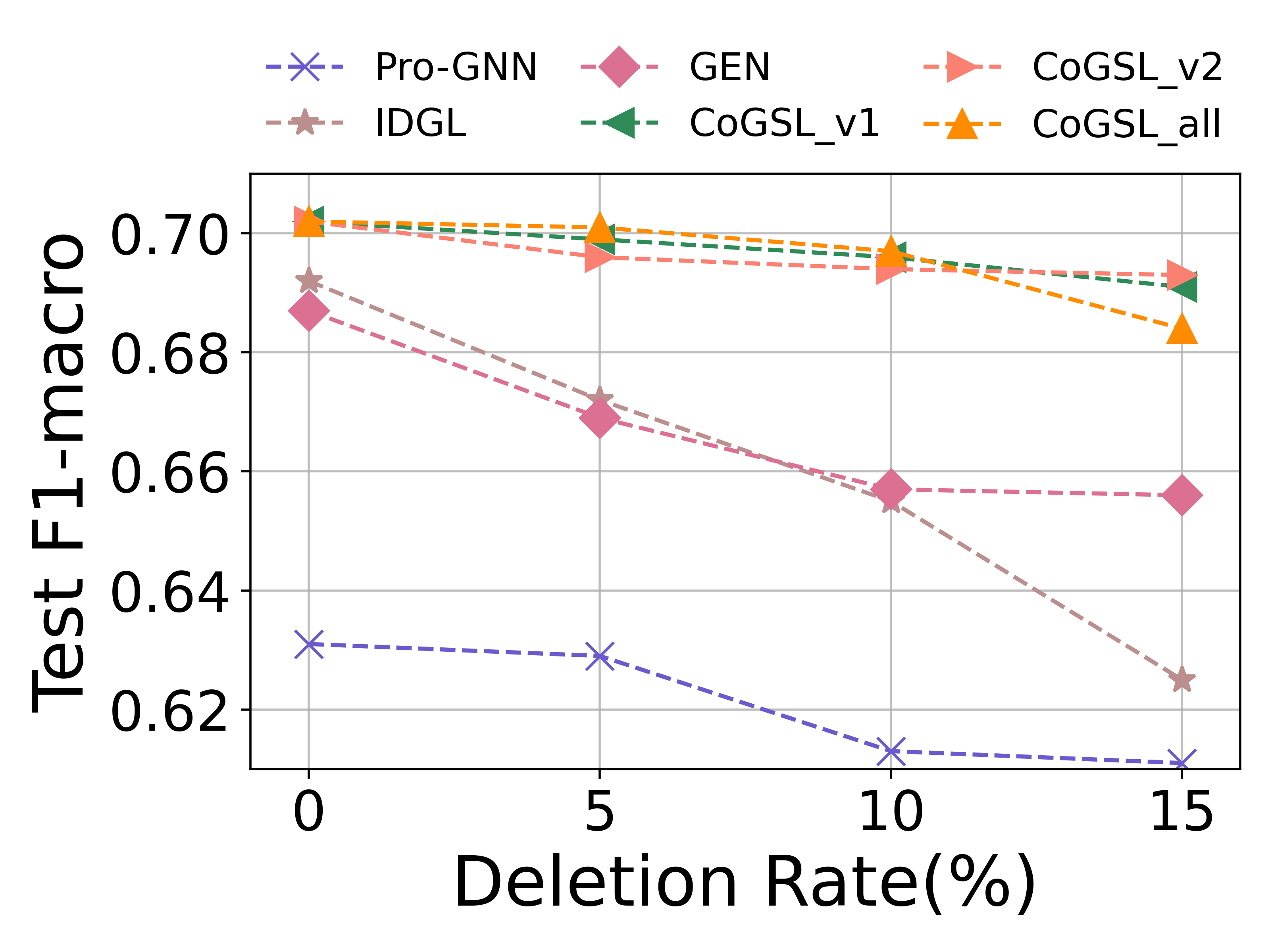}}
\quad\quad
\subfigure[Wiki-CS]{
\label{dele_wikics}
\includegraphics[scale=0.25]{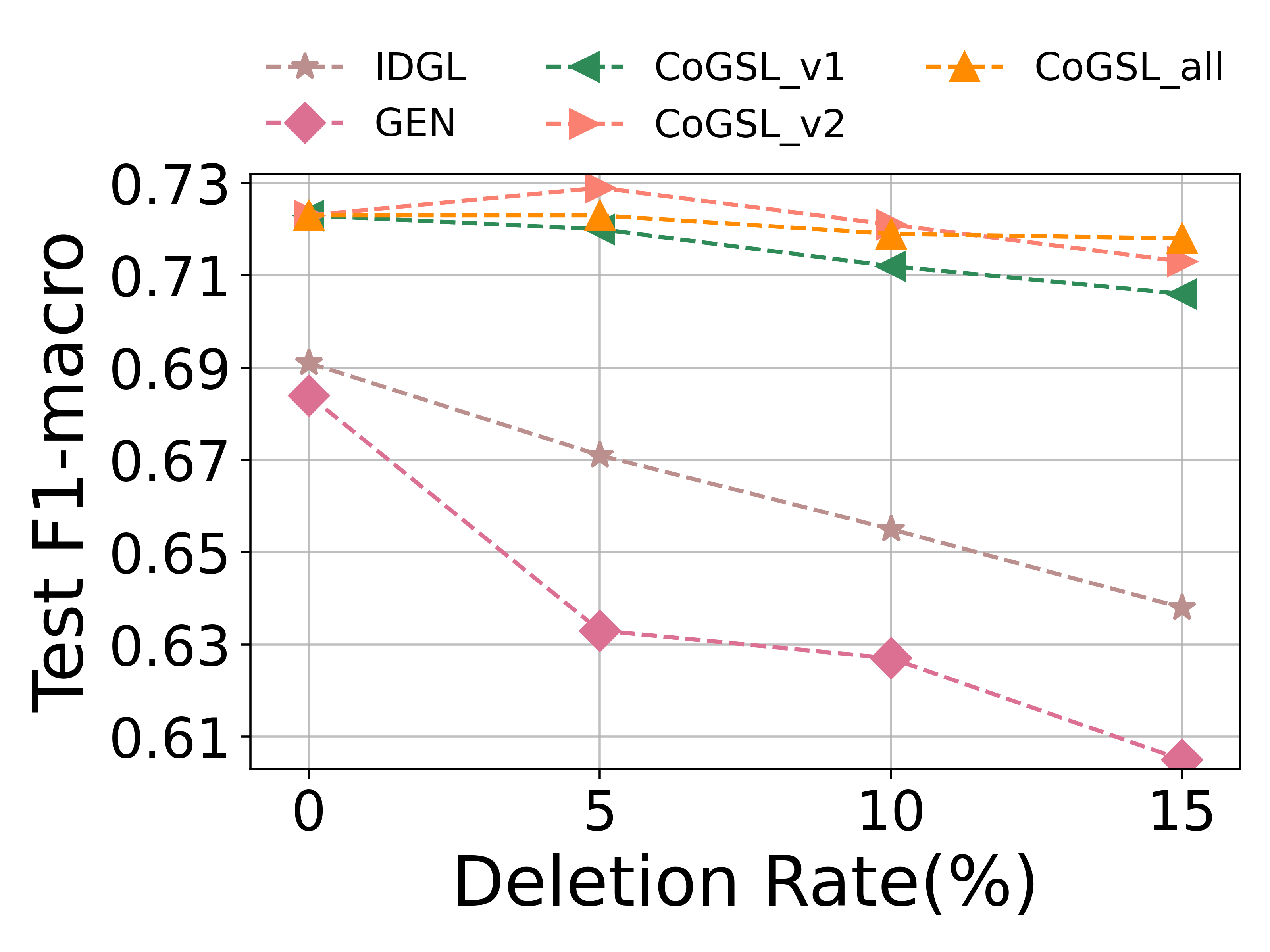}}
\caption{Results of different models under random edge deletion.}
\label{dele}
\end{figure*}

\begin{figure*}[tbp]
\centering
\subfigure[Cancer]{
\label{add_cancer}
\includegraphics[scale=0.25]{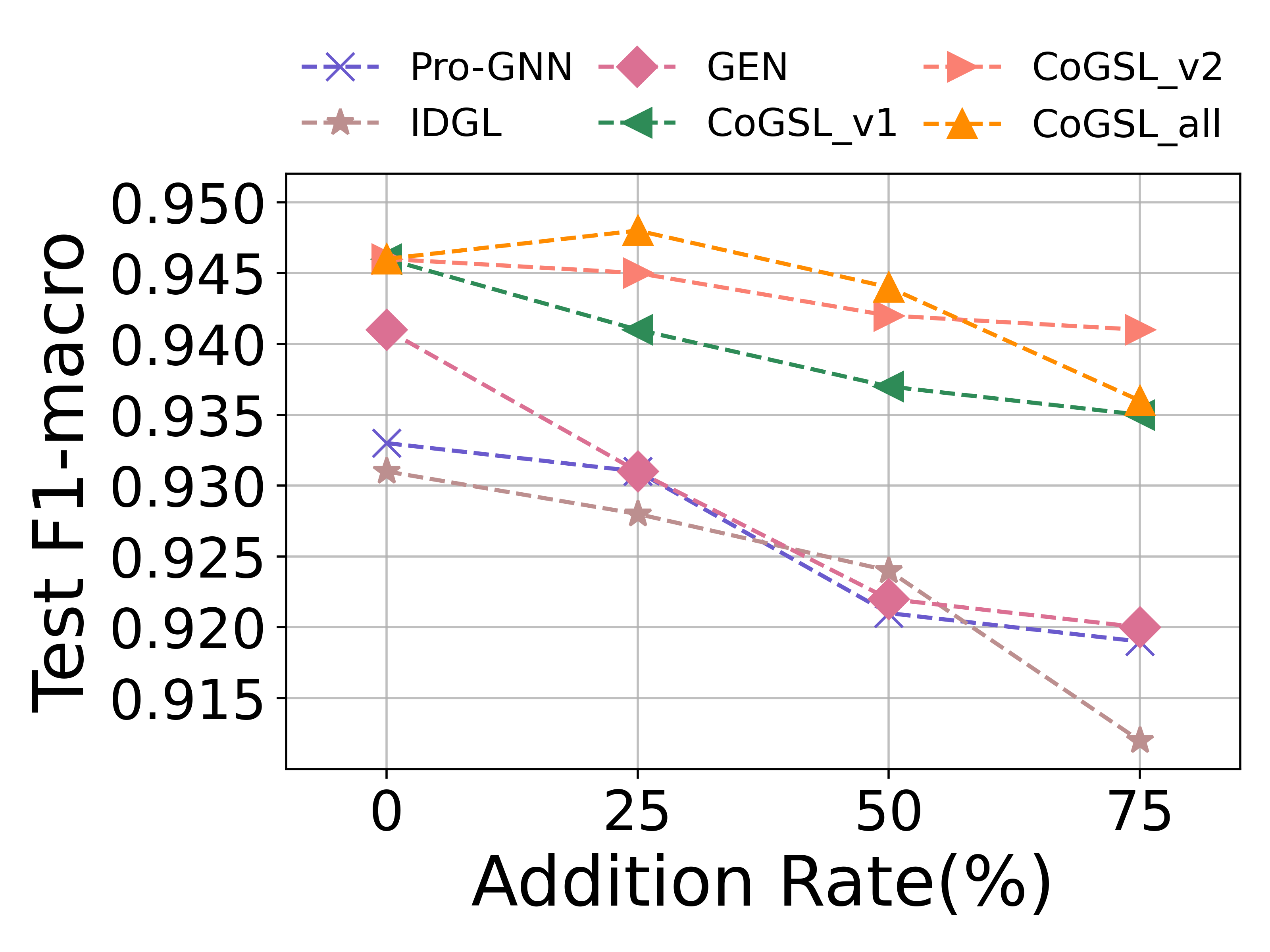}
}
\quad\quad
\subfigure[Citeseer]{
\label{add_citeseer}
\includegraphics[scale=0.25]{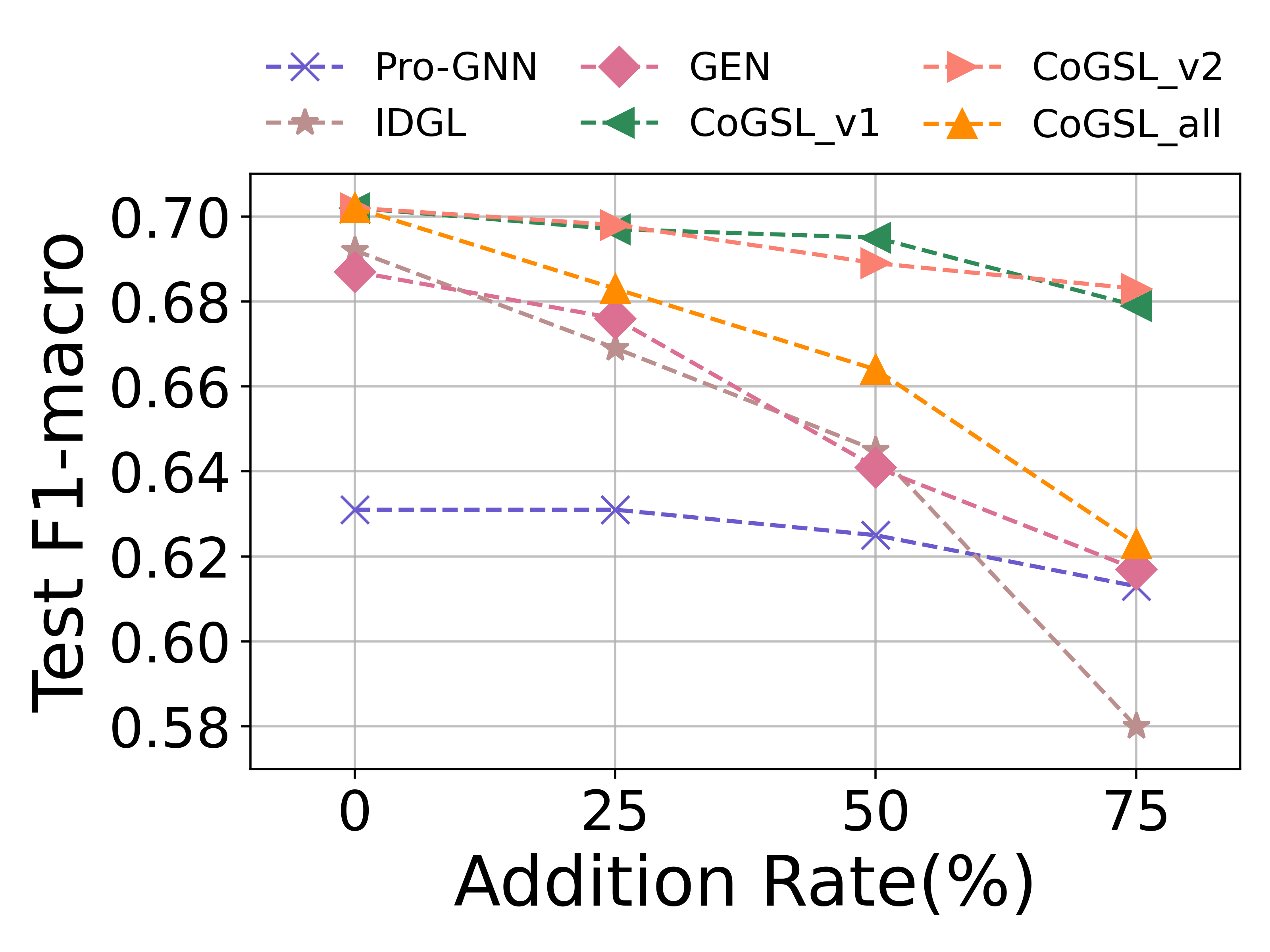}}
\quad\quad
\subfigure[Wiki-CS]{
\label{add_wikics}
\includegraphics[scale=0.25]{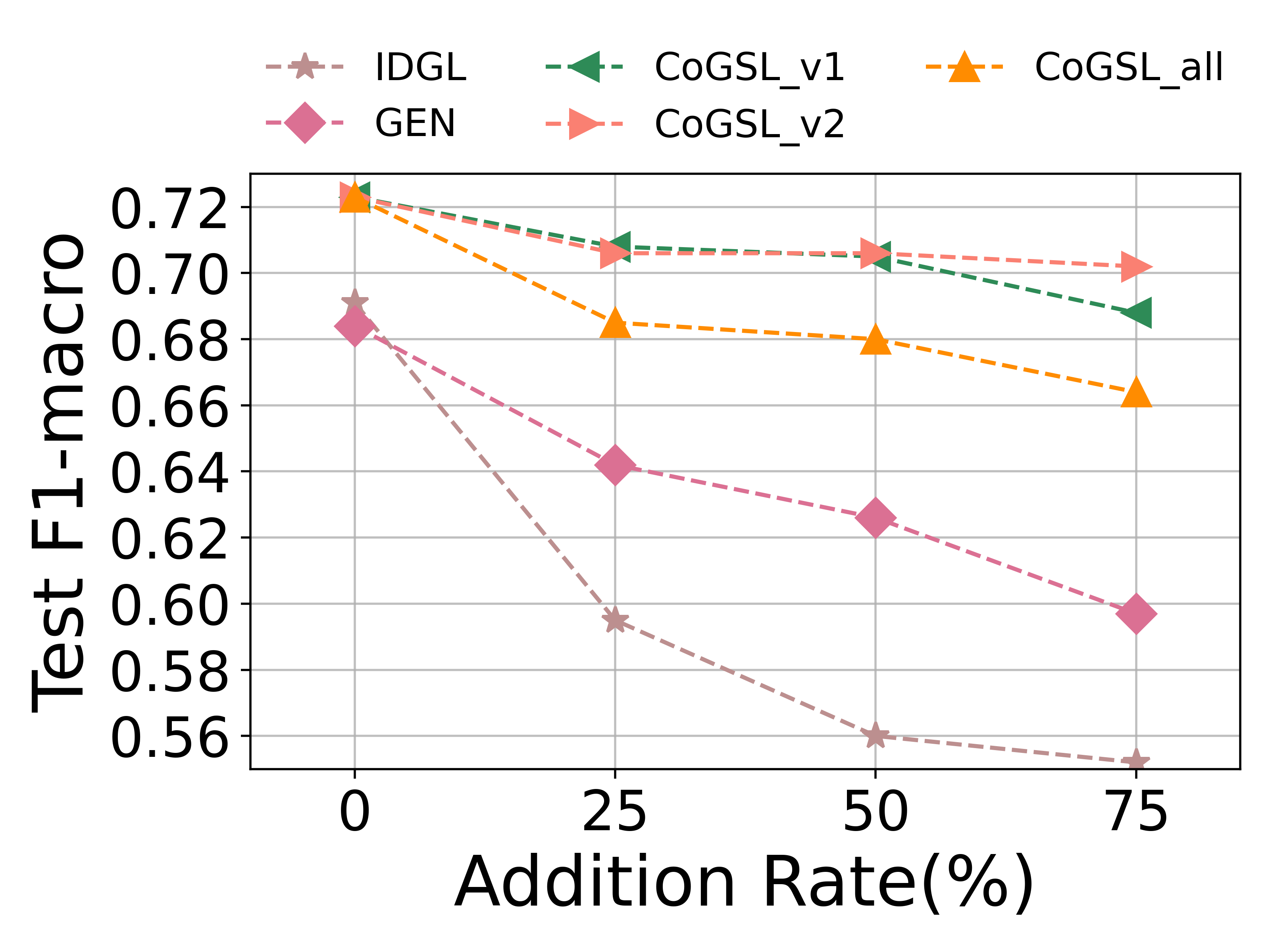}}
\caption{Results of different models under random edge addition.}
\label{add}
\end{figure*}

\textbf{Datasets}\quad We employ seven open datasets, including three academic networks (i.e., Citeseer \cite{gcn}, Wiki-CS \cite{wikics} and MS Academic \cite{ms}), three non-graph datasets (i.e., Wine, Breast Cancer (Cancer) and Digits) available in scikit-learn \cite{sklearn} and a blog graph Polblogs \cite{prognn}. The basic information about datasets is summarized in appendix~\ref{app_data}. Notice that for non-graph datasets, we construct a KNN graph as an initial adjacency matrix as in \cite{idgl}.


\noindent\textbf{Baselines}\quad We compare the proposed CoGSL with three categories of baselines: MI based unsupervised methods \{DGI \cite{dgi}, GCA \cite{gca}\}, classical GNN models \{GCN \cite{gcn}, GAT \cite{gat}, GraphSAGE \cite{graphsage}\} and three graph structure learning based methods \{Pro-GNN \cite{prognn}, IDGL \cite{idgl}, GEN \cite{gen}\}.

\noindent\textbf{Implementation Details}\quad For DGI and GCA, we firstly generate node embeddings, and then evaluate embeddings following the way stated in original papers. For three classical GNN models (i.e. GCN, GAT, GraphSAGE), we adopt the implementations from PyTorch Geometric library \cite{pyg}. For Pro-GNN, IDGL and GEN, we use the source codes provided by authors, and follow the settings in their original papers with carefully tune. For the proposed CoGSL, we use Glorot initialization \cite{glorot2010understanding} and Adam \cite{adam} optimizer. We carefully select two basic views for different datasets as two inputs, which are summarized in appendix~\ref{app_view}. We set the learning rate for classifiers $\Theta$ and MI estimator $\Phi$ as 0.01, and tune it for view estimator $\Omega$ from \{0.1, 0.01, 0.001\}. For combination coefficient $\mu$, we test ranging from \{0.1, 0.5, 1.0\}. We set $\epsilon$ as 0.1 and search on $\lambda$ from 0.1 to 0.9. Finally, we carefully select total iterations $T$ from \{100, 150, 200\}, and tune training epochs for each fold \{$\rho_{\Theta}, \rho_{\Phi}, \rho_{\Omega}$\} from \{1, 5, 10\}. The source code and datasets are publicly available on Github$\footnote{https://github.com/liun-online/CoGSL}$.

For fair comparisons, we set the hidden dimension as 16 and randomly run 10 times and report the average results for all methods. For Pro-GNN, IDGL, GEN and our CoGSL, we uniformly choose two-layer GCN as backbone to valuate the learnt structure. For the reproducibility, we report the related parameters in appendix~\ref{app_imp}.

\subsection{Node Classification}
\label{exp1}
In this section, we evaluate the proposed CoGSL on semi-supervised node classification. For different datasets, we follow the original splits on training set, validation set and test set. To more comprehensively evaluate our model, we use three common evaluation metrics, including F1-macro, F1-micro and AUC. The results are reported in Table \ref{fenlei}, where we randomly run 10 times and report the average results. The "-" symbol in Table \ref{fenlei} indicates that experiments could not be conducted due to memory issue. As can be seen, the proposed CoGSL generally outperforms all the other baselines on all datasets, which demonstrates that CoGSL can boost node classification in an effective way. The huge performance superiority of CoGSL over backbone GCN implies that view estimator and classifier are collaboratively optimized, and promote each other. In comparison with other GSL frameworks, our performance improvement illustrates that proposed principles are valid, and the learnt minimal sufficient structure with more effective information and less noise can offer a better solution. 

\subsection{Defense Performance}
Here, we aim to evaluate the robustness of various methods. To comprehensively conduct evaluation, we adopt three datasets with different scales, Cancer, Citeseer and Wiki-CS. We focus on comparing with GSL models, because these models can adjust the original structure, which makes them more robust than other GNNs. Specifically, we choose Pro-GNN as the representative of single-view based methods. And for multi-view based methods, IDGL and GEN are both selected.

\subsubsection{Attacks on edges}
\label{attack_edge}
To attack edges, we adopt random edge deletions or additions following \cite{lds, idgl}. Specifically, for edge deletions, we randomly remove 5\%, 10\%, 15\% of original edges, which retains the connectivity of attacked graph. For edge addition, we randomly inject fake edges into the graph by a small percentages of the number of original edges, i.e. 25\%, 50\%, 75\%. In view of that our CoGSL needs two inputs while other methods need one input, for a fair comparison, we deploy attacks on each of two inputs separately and on both of them together with the same percentages. We choose poisoning attack \cite{gib}, where we firstly generate attacked graphs and then use them to train models. All the experiments are conducted 10 times and we report the average accuracy. The results are plotted in Fig.~\ref{dele} and~\ref{add}. Notice that we do not conduct Pro-GNN on Wiki-CS because of time consuming (more than two weeks for a result). Besides, the curves of "CoGSL\_v1", "CoGSL\_v2" and "CoGSL\_all" mean the results that one of inputs of CoGSL is attacked and both of them are attacked, respectively. 

From the figures, CoGSL consistently outperforms all other baselines under different perturbation rates by a margin for three cases. We also find that as the perturbation rate increases, the margin becomes larger, which indicates that our model is more effective with violent attack. Besides, "CoGSL\_all" also performs competitive. Although both of its two inputs are attacked, "CoGSL\_all" still outperforms other baselines. 

\begin{table}[h]
  \caption{Quantitative results under feature attack.}
  \label{feat_attack}
  \resizebox{0.3\textwidth}{!}{
  \begin{tabular}{c|c|cccc}
    \bottomrule
    Datasets & F1-macro & Pro-GNN & IDGL & GEN & CoGSLL\\
    \bottomrule
    \multirow{4}{*}{Cancer}
    &0.0&93.3&93.1&94.1&\textbf{94.6}\\
    &0.1&92.9&91.5&92.9&\textbf{94.2}\\
    &0.3&92.6&90.5&91.9&\textbf{93.6}\\
    &0.5&92.2&90.2&90.9&\textbf{93.4}\\
    \hline
    \multirow{4}{*}{Citeseer}
    &0.0&63.1&69.2&68.7&\textbf{70.2}\\
    &0.1&55.5&64.1&65.3&\textbf{67.8}\\
    &0.3&44.1&22.6&36.1&\textbf{49.1}\\
    &0.5&36.8&23.3&29.4&\textbf{43.5}\\
    \hline
    \multirow{4}{*}{Wiki-CS}
    &0.0&-&69.1&68.4&\textbf{72.3}\\
    &0.1&-&63.6&46.8&\textbf{70.4}\\
    &0.3&-&41.6&24.2&\textbf{46.2}\\
    &0.5&-&12.5&18.5&\textbf{24.2}\\
    \bottomrule
  \end{tabular}}
\end{table}
\subsubsection{Attacks on features}
To attack feature, we add independent Gaussian noise to features as in \cite{gib}. Specifically, we firstly sample a noise matrix $M_{noise}\in\mathbf{R}^{N\times D}$, where each entry is sampled from $N(0, 1)$. Then, we calculate reference amplitude $r$, which is the mean of maximal value of each node's feature. We add $\aleph \cdot r\cdot M_{noise}$ to original feature matrix $X$, and get the attacked feature matrix $X_{noise}$, where $\aleph\in\{0.1, 0.3, 0.5\}$ is the noise ratio. We also conduct poisoning settings and report the results in Table \ref{feat_attack}, where the results of Pro-GNN on Wiki-CS are not reported for the same reason in section~\ref{attack_edge}. Again, CoGSL consistently outperforms all other baselines and successfully resists attacks on features. Together with observations from~\ref{attack_edge}, we can conclude that CoGSL can approach the minimal sufficient structure, so it is able to defend attacks from edges and features.

\subsection{Model Analysis}
\subsubsection{Analysis of view estimator}
\begin{figure}[h]
\centering
\subfigure[Digits]{
\label{xiao_digits}
\includegraphics[scale=0.2]{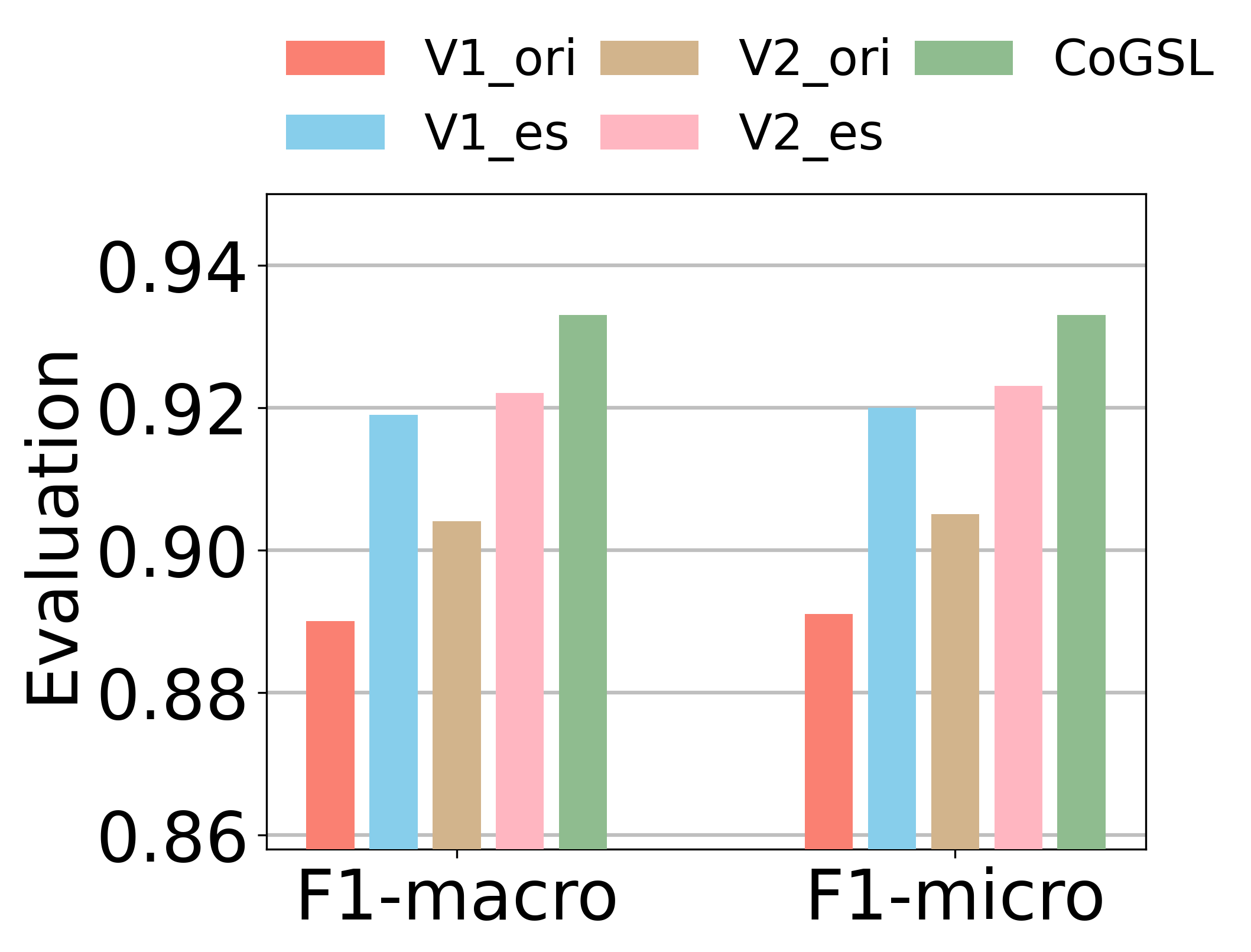}}
\quad\quad
\subfigure[Citeseer]{
\label{xiao_citeseer}
\includegraphics[scale=0.2]{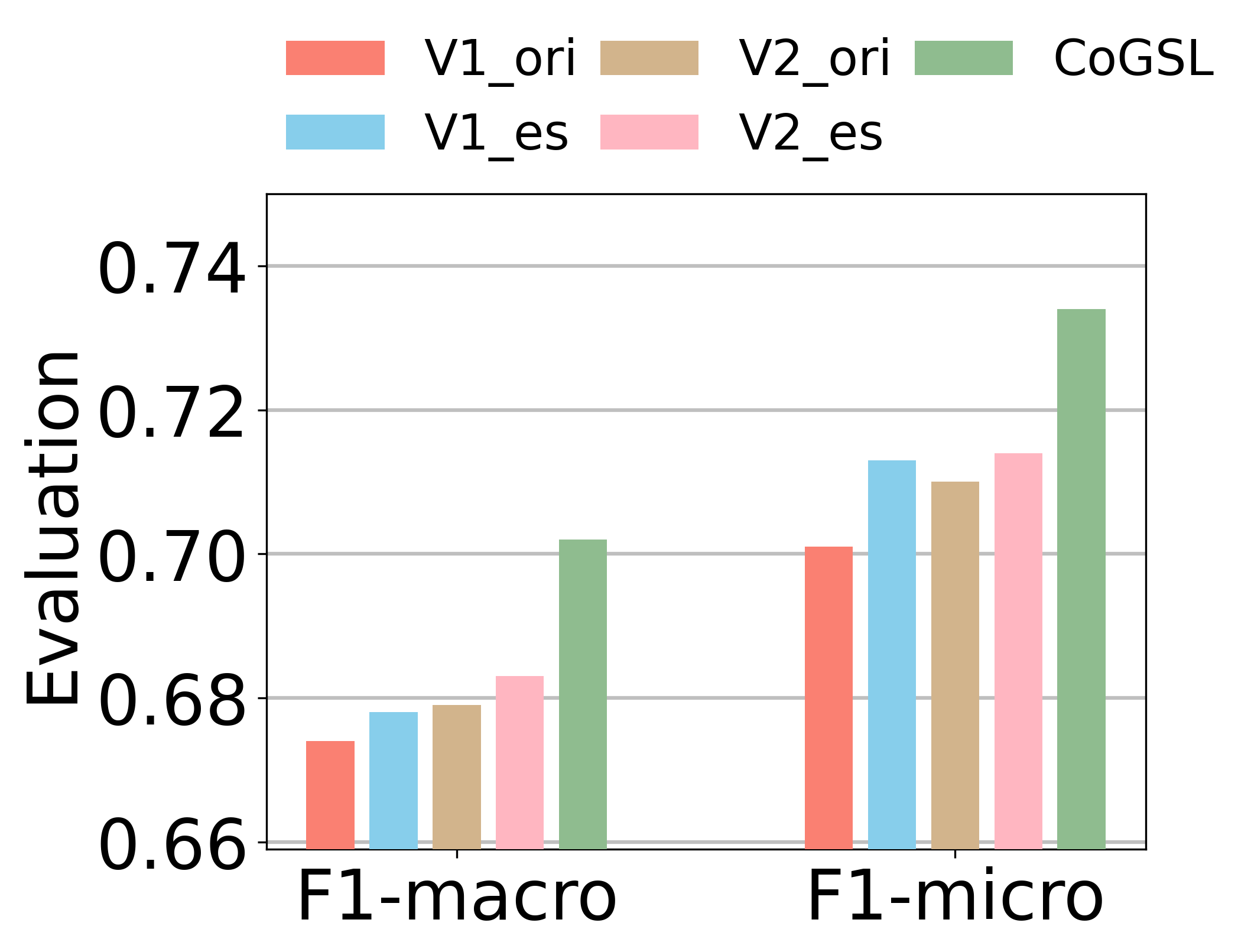}
}
\caption{Test on the effectiveness of view estimator.}
\label{ablation}
\end{figure}

Our model involves two basic views as inputs, each of which will be reestimated with the view estimator. To evaluate the effectiveness of view estimator, we firstly train the model, and pick two final estimated views. After that, we compare the performance of two original views, two final estimated views and the final view. We conduct comparison on Citeseer and Digits, and the results are given in Fig.~\ref{ablation}, where $V1\_ori$ and $V2\_ori$ mean two original views, and $V1\_es$ and $V2\_es$ are two estimated views. We can see that all estimated views gain an improvement over corresponding original views, which indicates the effectiveness of view estimator. Moreover, CoGSL always outperforms the estimated views, and this proves the reliability of adaptive fusion and following optimization principles.

\subsubsection{Analysis of adaptive fusion}
We propose an adaptive fusion mechanism, which assigns weights to two estimated views based on the confidence on them for each node as eq.~\eqref{prediction}-\eqref{end} in section~\ref{View Fusion}. To verify the validation of this part, we design two more baselines. One is to simply average two estimated views as the final view. The other is to use attention mechanism to fuse them, where we adopt a channel attention layer in \cite{hgsl}. We test on Citeseer and Digits and show the results in Table~\ref{adpative}, where "Adaption" refers to adaptive fusion we introduce. We can see that our newly proposed adaptive fusion is the best behaved of three ways. Also, we notice that "Average" behaves better than "Attention", and we think this is because "Attention" fusion involves some new parameters, which increases the complexity of model and brings the risk of over-fitting.

\begin{table}[h]
  \caption{Quantitative results on different fusions.}
  \label{adpative}
  \resizebox{0.35\textwidth}{!}{
  \begin{tabular}{c|ccc|ccc}
    \bottomrule
     & \multicolumn{3}{c|}{Digits}&\multicolumn{3}{c}{Citeseer}\\
    \bottomrule
    Fusion&F1-ma&F1-mi&AUC&F1-ma&F1-mi&AUC\\
    \hline
    Average&93.0&93.0&99.5&69.6&72.8&90.8\\
    Attention&92.9&93.0&99.6&69.4&72.7&91.2\\
    \hline
    \textbf{Adaption}&\textbf{93.3}&\textbf{93.3}&\textbf{99.6}&\textbf{70.2}&\textbf{73.4}&\textbf{91.4}\\
    \bottomrule
  \end{tabular}}
\end{table}

\subsubsection{Analysis of MI}
\begin{figure}[h]
\centering
\subfigure[Digits]{
\label{mi_digits}
\includegraphics[scale=0.2]{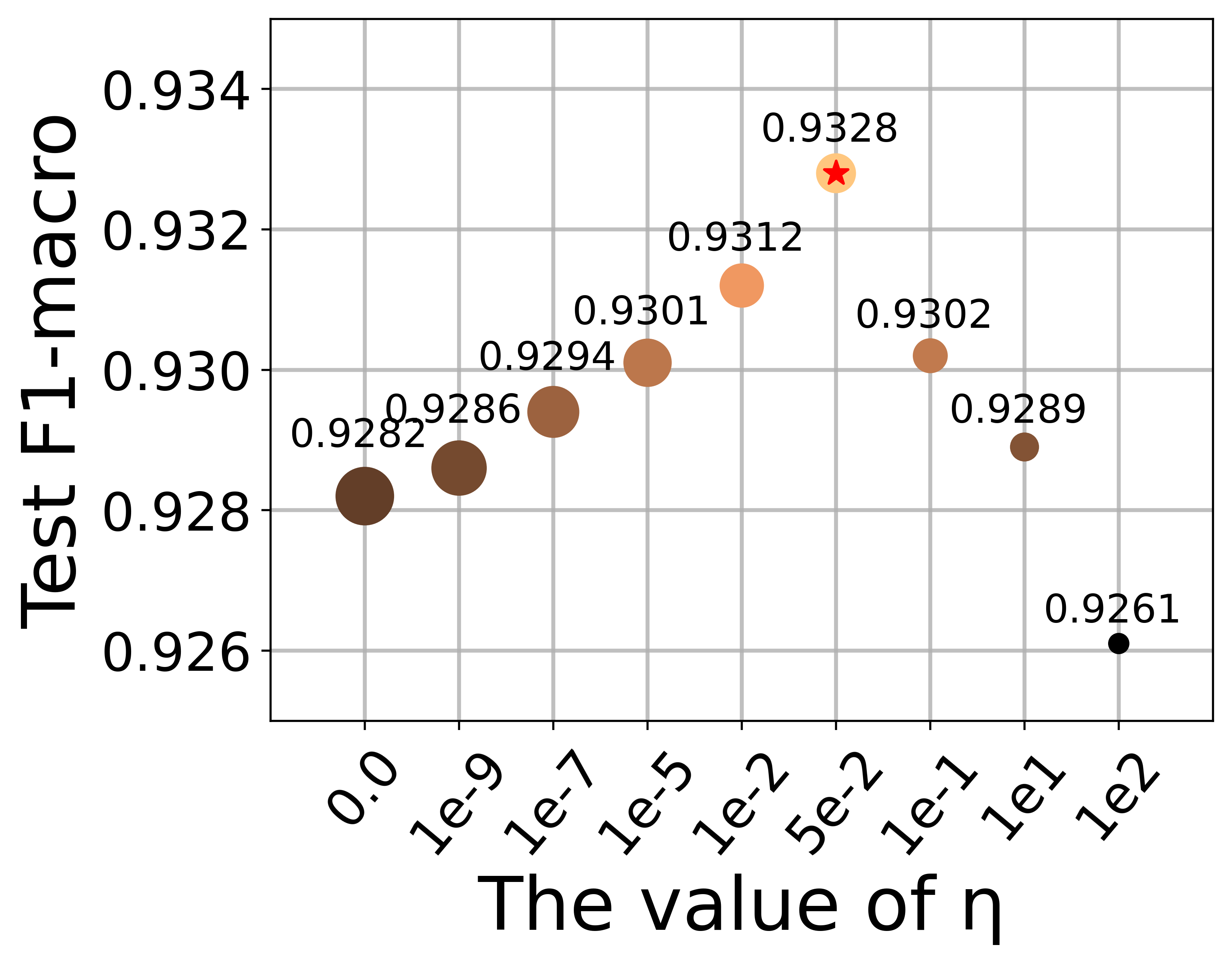}
}
\ 
\subfigure[Citeseer]{
\label{mi_citeseer}
\quad
\includegraphics[scale=0.2]{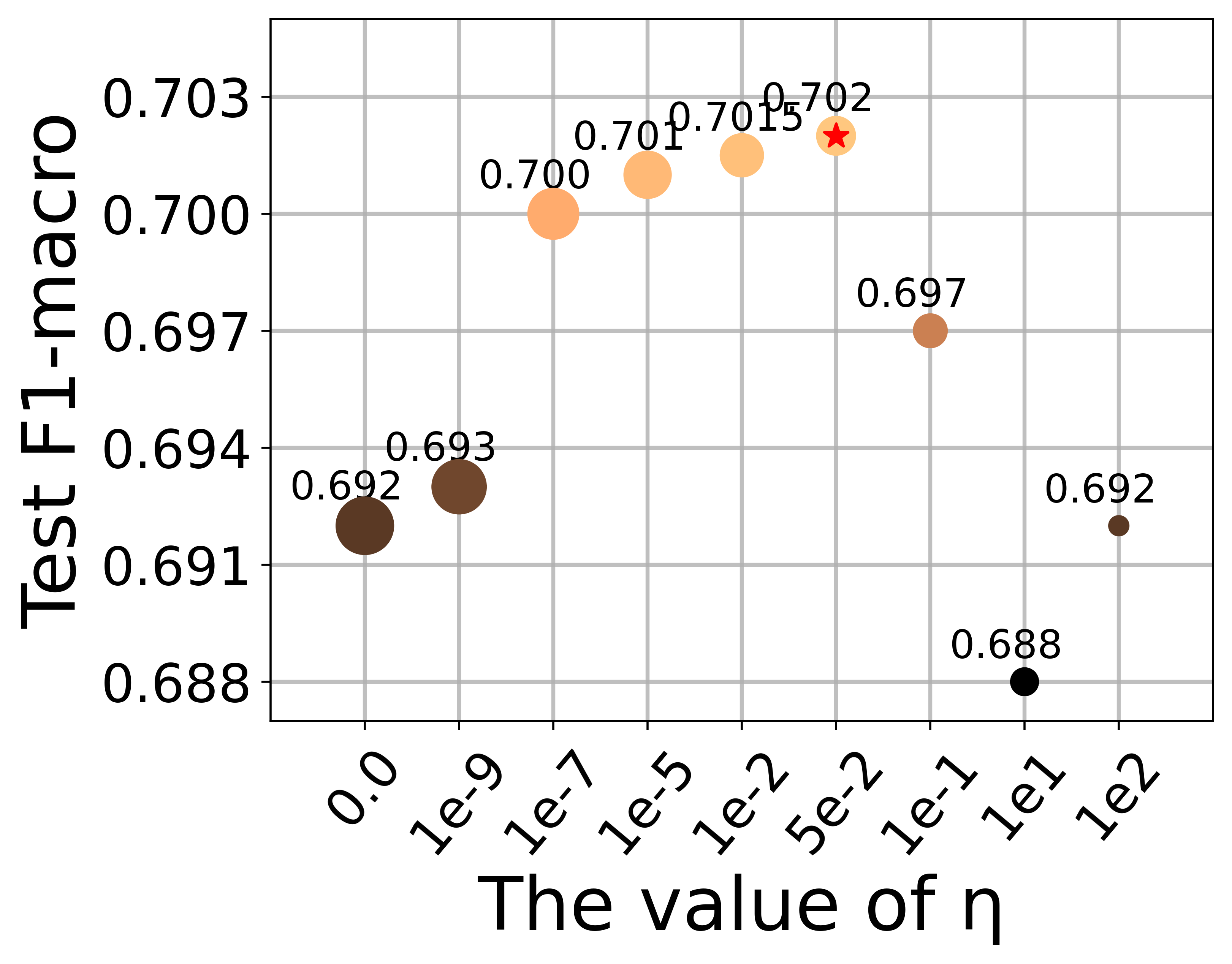}
}
\caption{The investigation of change of MI.}
\label{test_mi}
\end{figure}

We need to constrain the MI between views are neither too weak or too strong, so that the final view contain concise information, no more and no less.
We notice that as a balance parameter, $\eta$ in eq.~\eqref{trian_omega} well controls the effect of MI loss. If $\eta$ increases, MI between views is heavily constrained, and vice versa. So, we investigate the change of $\eta$ to substitute the change of real MI between views, and the results are shown in Fig.~\ref{test_mi}, where we report the results on Citeseer and Digits. In this figure, the area of each point means relative size of MI between views. The shallower the color of point is, the better the performance is. And the best point is marked with a red star. We observe that the optimal point is a medium value, neither a too strong or a too weak constraint. Especially, when $\eta$ equals to zero, we mimic the situation of general GSL methods, and we can see that the results are not very good in this case. It implies that restricting MI between views is necessary.

\begin{figure}[h]
\centering
\subfigure[Citeseer: adjacency matrix]{
\label{citeseer_adj}
\includegraphics[scale=0.2]{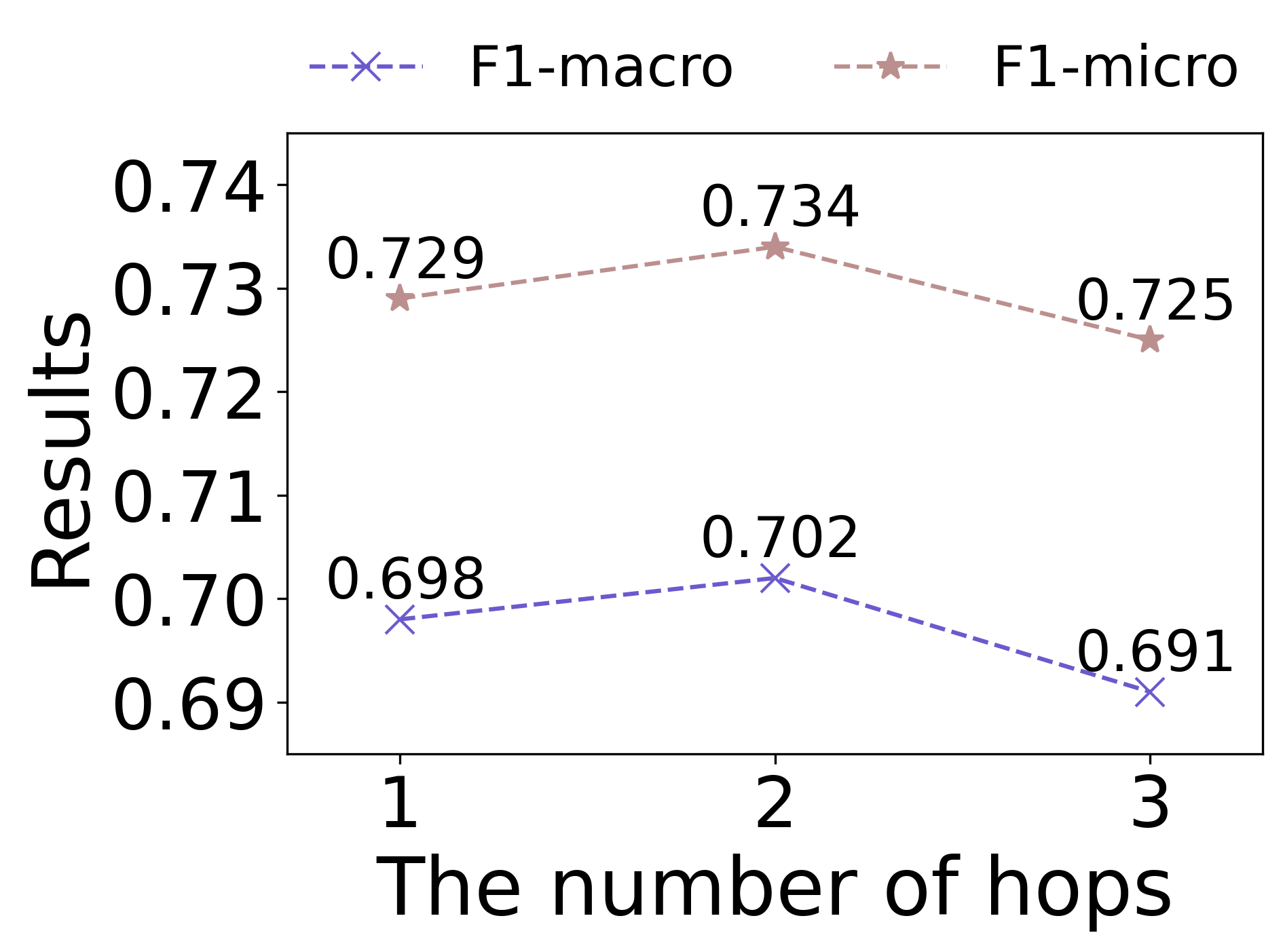}
}
\quad
\subfigure[Citeseer: diffusion matrix]{
\label{citeseer_diff}
\includegraphics[scale=0.2]{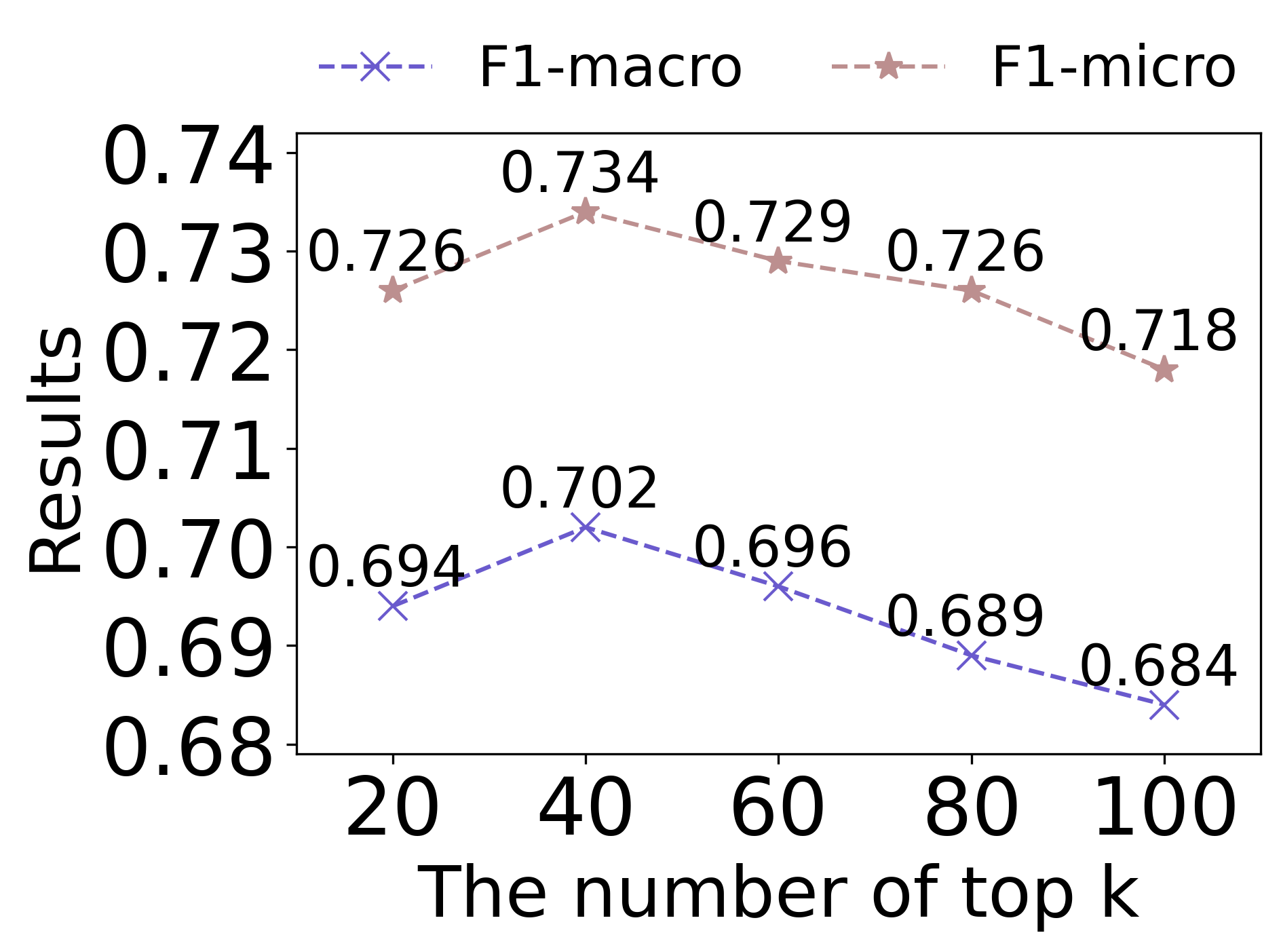}
}
\ 
\subfigure[Digits: adjacency matrix]{
\label{citeseer_diff}
\includegraphics[scale=0.2]{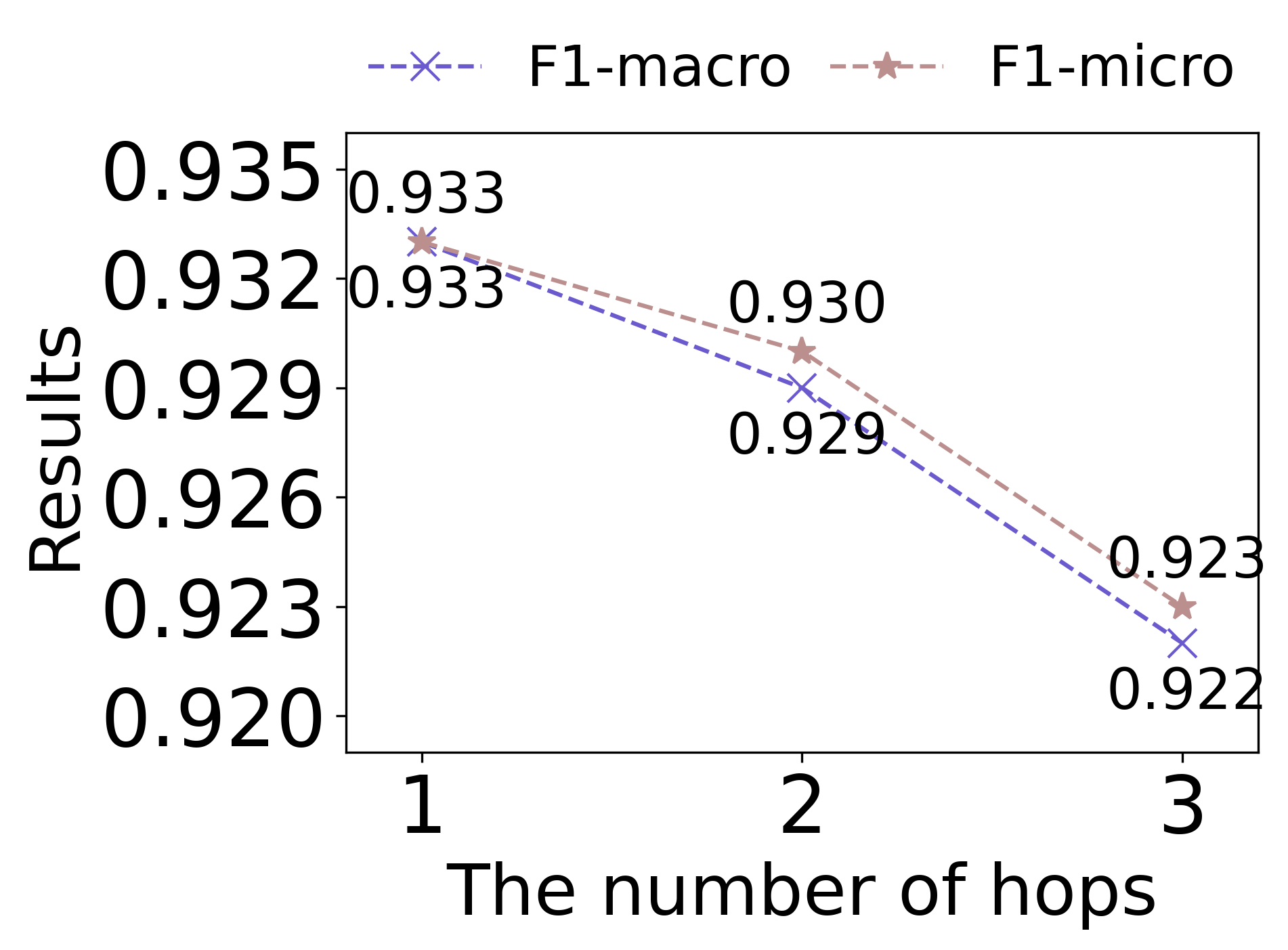}
}
\quad
\subfigure[Digits: diffusion matrix]{
\label{citeseer_diff}
\includegraphics[scale=0.2]{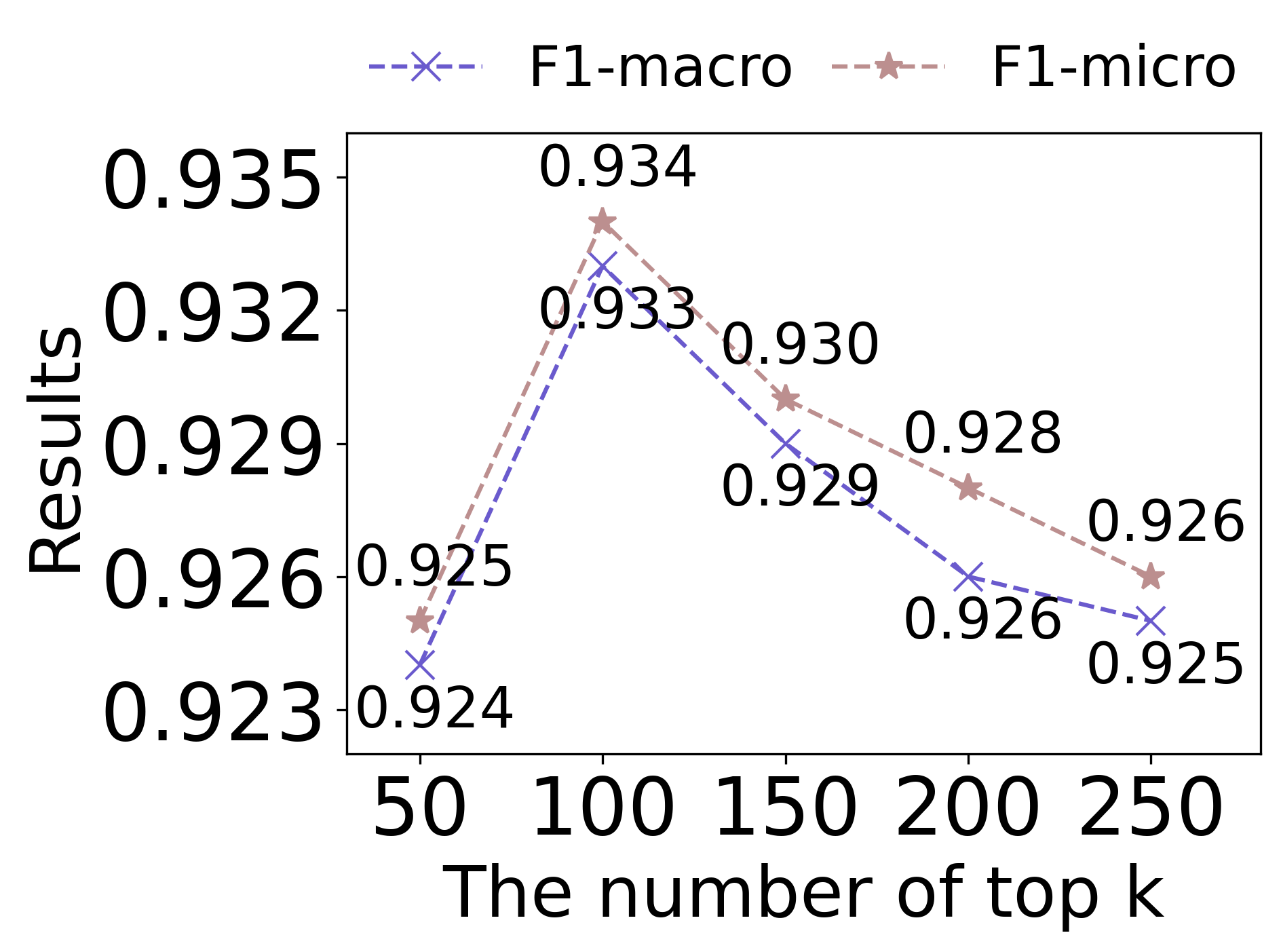}
}
\caption{Impact of hyper-parameter scope.}
\label{scope}
\end{figure}

\subsubsection{Analysis of hyper-parameter}
In this section, we explore the sensitivity of h and k on Citeseer and Digits, introduced in section~\ref{Estimator}. As shown in appendix~\ref{app_view}, the input views of Citeseer and Digits are both adjacency matrix and diffusion matrix, plotted in Fig.~\ref{scope}. For Digits, the optimal $h$ of adjacency matrix is 1-hop, and the optimal $k$ of diffusion matrix is top 100. Similarly, for Citeseer, the optimal points are 2-hop and top 40. We can see that a proper estimation scope is indispensable. If the scope is too small, some important structures are neglected. However if the scope is too large, we can not distinguish the right connections effectively.


\section{conclusion}
In this paper, we theoretically study how to estimate a minimal sufficient structure in GSL problem. We prove that if the performances of basic views and final view are maintained, the mutual information between every two views should be minimized simultaneously, so that the learnt final view tends to be minimal sufficient structure. With this theory, we propose CoGSL, a framework to learn a compact graph structure by compressing mutual information. Extensive experimental results, under clean and attacked conditions, are conducted to verify the effectiveness and robustness of CoGSL.

\begin{acks}
This work is supported in part by the National Natural Science Foundation of China (No. U20B2045, 62192784, 62172052, 61772082, 62002029). It is also supported in part by the National Natural Science Foundation of China (62172052, U1936104) and The Fundamental Research Funds for the Central Universities 2021RC28. 
\end{acks}

\bibliographystyle{ACM-Reference-Format}
\bibliography{sample-base}

\appendix
\newpage
\section{Details on Experimental Setup}
\label{DDDD}
In this section, for the reproducibility, we provide some basic information about baselines and datasets. The implementation details, including the detailed hyper-parameter values, are also provided.

\subsection{Datasets}
\label{app_data}

Table~\ref{statistics} shows the statistics of seven datasets used in our experiments.
\begin{table}[h]
  \caption{The statistics of the datasets}
  \label{statistics}
  \resizebox{0.45\textwidth}{!}{
  \begin{tabular}{c|ccccc}
    \bottomrule
        Dataset & Nodes & Edges & Classes & Features & Train/Val/Test\\
    \bottomrule
    Wine & 178 & 3560 & 3 & 13 & 10/20/148 \\
    Cancer & 569 & 22760 & 2 & 30 & 10/20/539 \\
    Digits & 1797 & 43128 & 10 & 64 & 50/100/1647 \\
    \bottomrule
    Polblogs & 1222 & 33428 & 2 & 1490 & 121/123/978 \\
    Citeseer & 3327 & 9228 & 6 & 3703 & 120/500/1000 \\
    Wiki-CS & 11701 & 291039 & 10 & 300 & 200/500/1000 \\
    MS Academic & 18333 & 163788 & 15 & 6805 & 300/500/1000 \\
    \bottomrule
\end{tabular}}
\end{table}

These seven datasets used in experiments can be found in these URLs:
\begin{itemize}
    \item Wine, Breast Cancer and Digits: \url{https://scikit-learn.org/stable/modules/classes.html#module-sklearn.datasets}
    \item Polblogs: \url{https://github.com/ChandlerBang/Pro-GNN}
    \item Citeseer: \url{https://github.com/tkipf/gcn}
    \item Wiki-CS: \url{https://github.com/pmernyei/wiki-cs-dataset}
    \item MS Academic: \url{https://github.com/klicperajo/ppnp}
\end{itemize}

\subsection{Baselines}
\label{app_base}
The publicly available implementations of baselines can be found at the following URLs:

\begin{itemize}
    \item DGI: \url{https://github.com/PetarV-/DGI}
    \item GCA: \url{https://github.com/CRIPAC-DIG/GCA}
    \item GCN, GAT and GraphSAGE: \url{https://pytorch-geometric.readthedocs.io/en/latest/}
    \item LDS: \url{https://github.com/lucfra/LDS-GNN}
    \item Pro-GNN: \url{https://github.com/ChandlerBang/Pro-GNN}
    \item IDGL: \url{https://github.com/hugochan/IDGL}
    \item GEN: \url{https://github.com/BUPT-GAMMA/Graph-Structure-Estimation-Neural-Networks}
\end{itemize}

\subsection{The selected input views}
\label{app_view}
Table~\ref{selected} shows the basic views we select for different datasets.
\begin{table}[h]
  \caption{The selected views for different datasets}
  \label{selected}
  \resizebox{0.5\textwidth}{!}{
  \begin{tabular}{c|ccccccc}
    \bottomrule
        Candidate & Wine & Cancer & Digits & Polblogs & Citeseer & Wiki-CS & MS Academic\\
    \bottomrule
        Adjacency matrix ($A$)& & & $\surd$ & $\surd$ & $\surd$ & $\surd$ & $\surd$ \\
    \hline
        Diffusion matrix ($S$)& $\surd$ & $\surd$ & $\surd$ & $\surd$ & $\surd$ & &\\
    \hline
        KNN graph ($K$)& $\surd$ & $\surd$ & & & &&\\
    \hline
        Subgraph ($A_{sub}$)& & & & & & $\surd$ & $\surd$ \\
    \bottomrule
\end{tabular}}
\end{table}

\subsection{Hyperparameters Settings}
\label{app_imp}
We implement CoGSL in PyTorch, and list some important parameter values used in our model in Table~\ref{para}. In this table, \textit{ve\_lr} is the learning rate of view estimator, and \textit{ve\_drop} is the dropout used in estimating basic views. Notice that "-" of $B$ indicates that we use all of nodes to calculate InfoNCE loss.

\begin{table}[h]
  \caption{The values of parameter used in CoGSL.}
  \label{para}
  \resizebox{0.5\textwidth}{!}{
  \begin{tabular}{c||ccccccccc}
        \bottomrule
       Dataset & \textit{ve\_lr} & \textit{ve\_drop} & $T$ & $\rho_{\Theta}$ & $\rho_{\Phi}$ & $\rho_{\Omega}$ & $B$ & $\epsilon$ & $\lambda$ \\
        \bottomrule
       Wine & 0.001& 0.8& 100& 1& 5& 1& -&0.1 &0.5\\
       \hline
       Cancer & 0.1& 0.5& 150& 1& 5& 1& -&0.1 &0.9\\
       \hline
       Digits & 0.01& 0.5& 200& 10& 10& 1&- &0.1 &0.5\\
       \hline
       Polblogs & 0.1& 0.8& 150& 5& 5& 1& -& 0.1&0.1\\
       \hline
       Citeseer & 0.001& 0.2& 200& 5& 10& 5&- & 0.1&0.5\\
       \hline
       Wiki-CS & 0.01& 0.2& 200& 1& 5& 1& 1000& 0.1&0.1\\
       \hline
       MS Academic &0.0001 &0.8 &200 &15 &10 &1 &1000 &1.0 &0.2 \\
        \bottomrule
  \end{tabular}
  }
\end{table}

\section{Three-fold Optimization}
\label{alg}
In this section, we detail the process of three-fold optimization, shown in Algorithm 1.
\begin{algorithm}[htbp]
\caption{The CoGSL Algorithm}
\LinesNumbered
\SetKwInOut{Input}{\textbf{Input}}
\SetKwInOut{Params}{\textbf{Params}}
\SetKwInOut{Output}{\textbf{Output}}
\Input{Basic views \{$V_1$, $V_2$\}, feature matrix $\mathbf{X}$, labels $\mathcal{Y}_L$}
\Params{$B$, total iterations $T$, \\
training epochs for each fold \{$\rho_{\Theta}, \rho_{\Phi}, \rho_{\Omega}$\}}
\Output{final view $V^{\star}$, GCN parameters $\Theta$}
\BlankLine
Initialize $\Theta$, $\Phi$ and $\Omega$\;
\For{$i=1$ to $T$}{
\For{$j=1$ to $\rho_{\Omega}$}{
$\%\ View\ estimator\ Training$ \\
Estimate \{$V_1$, $V_2$\} as \{$V^1_{es}$, $V^2_{es}$\} with eq~\eqref{view_1}-~\eqref{view_2}\;
Adaptively fuse $V_1$ and $V_2$ into $V^{\star}$\;
Update $\Omega$ with eq.~\eqref{trian_omega}\;
}
Get \{$V^1_{es}$, $V^2_{es}, V^{\star}$\} after view estimating and fusion\;
\For{$k=1$ to $\rho_{\Theta}$}{
$\%\ Classifiers\ Training$ \\
Calculating predictions with eq.~\eqref{prediction} and~\eqref{prediction1}\;
Update $\Theta$ with eq.~\eqref{trian_theta}\;
}
\For{$l=1$ to $\rho_{\Phi}$}{
$\%\ MI\ estimator\ Training$ \\
Randomly sample B nodes to calculate eq.~\eqref{contra_loss}\;
Update $\Phi$ by minimizing eq.~\eqref{mi_loss}\;
}

}
\Return $V^{\star}$ and $\Theta$\;
\end{algorithm}

\end{document}